\documentclass[lettersize,journal]{IEEEtran}
\usepackage{amsmath,amsfonts}
\usepackage{algorithm}
\usepackage{array}
\usepackage{textcomp}
\usepackage{stfloats}
\usepackage{url}
\usepackage{verbatim}
\usepackage{graphicx}
\usepackage{cite}
\usepackage[caption=false,font=scriptsize,labelfont=sf,textfont=sf]{subfig}
\usepackage{color}
\usepackage{bm}
\usepackage{makecell}
\usepackage{pifont}
\usepackage{xcolor}
\usepackage{algpseudocode}
\usepackage{multirow}
\newcommand{\bzero}{\bm{0}}
\newcommand{\bI}{\bm{I}}

\usepackage{booktabs}
\usepackage{colortbl}
\usepackage{amssymb}

\newcommand{\second}[1]{\textcolor{blue}{\underline{#1}}}
\newcommand{\best}[1]{\textcolor{red}{\textbf{#1}}}
\usepackage{colortbl}
\newcommand{\mygraycell}{\cellcolor[gray]{0.9}}
\usepackage{hyperref}
\usepackage{siunitx}
\begin{document}

\title{LatentINDIGO: An INN-Guided Latent Diffusion Algorithm for Image Restoration}

\author{Di You,~\IEEEmembership{Graduate Student Member,~IEEE}, Daniel Siromani,~\IEEEmembership{Graduate Student Member,~IEEE}, and Pier~Luigi~Dragotti,~\IEEEmembership{Fellow,~IEEE}
        % <-this % stops a space
\thanks{Di You, Daniel Siromani, and Pier~Luigi~Dragotti are with the Department
of Electrical and Electronic Engineering, Imperial College London, United
Kingdom (e-mail: di.you22@imperial.ac.uk; d.siromani23@imperial.ac.uk; p.dragotti@imperial.ac.uk).}% <-this % stops a space
}



\maketitle
\begin{abstract}
There is a growing interest in the use of latent diffusion models (LDMs) for image restoration (IR) tasks due to their ability to model effectively the distribution of natural images. While significant progress has been made, there are still key challenges that need to be addressed. First, many approaches depend on a predefined degradation operator, making them ill-suited for complex or unknown degradations that deviate from standard analytical models. Second, many methods struggle to provide a stable guidance in the latent space and finally most methods convert latent representations back to the pixel domain for guidance at every sampling iteration, which significantly increases computational and memory overhead. To overcome these limitations, we introduce a wavelet-inspired invertible neural network (INN) that simulates degradations through a forward transform and reconstructs lost details via the inverse transform. We further integrate this design into a latent diffusion pipeline through two proposed approaches: LatentINDIGO-PixelINN, which operates in the pixel domain, and LatentINDIGO-LatentINN, which stays fully in the latent space to reduce complexity. Both approaches alternate between updating intermediate latent variables under the guidance of our INN and refining the INN forward model to handle unknown degradations. In addition, a regularization step preserves the proximity of latent variables to the natural image manifold. Experiments demonstrate that our algorithm achieves state-of-the-art performance on synthetic and real-world low-quality images, and can be readily adapted to arbitrary output sizes.
\end{abstract}

\section{Introduction}
\label{sec:intro}

\IEEEPARstart{I}{mage} restoration (IR) is a classic inverse problem aiming to recover high-quality images from their noisy and degraded measurements.
In a typical restoration problem, one observes $\bm{y} =\mathcal{H}(\bm{x},\bm{n})$, where $\bm{y}$ is the degraded and noisy version of the original image $\bm{x}$ and $\bm{n}$ is some noise. The degradation process $\mathcal{H}$ can be linear or non-linear. To deal with real-world degraded images, Blind Image Restoration (BIR) refers to the case where one aims to reconstruct the original image without prior knowledge of the degradation process, and this is a situation that happens frequently in many real-world scenarios.

\begin{figure}[!tp]\footnotesize 
	\centering
\hspace{-0.2cm}
\begin{tabular}{c@{\extracolsep{0.05em}}c@{\extracolsep{0.05em}}c@{\extracolsep{0.05em}}c@{\extracolsep{0.05em}}c}
            &\includegraphics[width=0.115\textwidth]{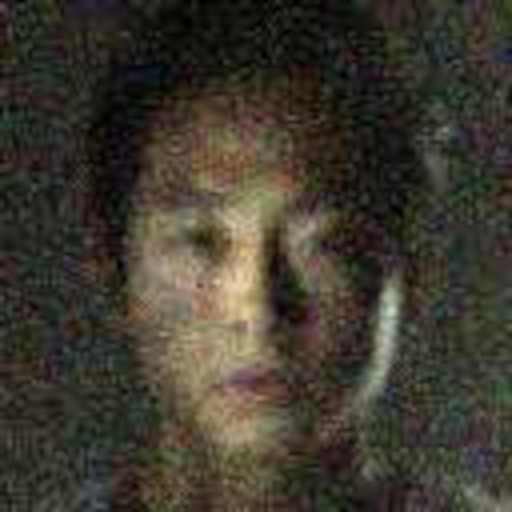}~
    &\includegraphics[width=0.115\textwidth]{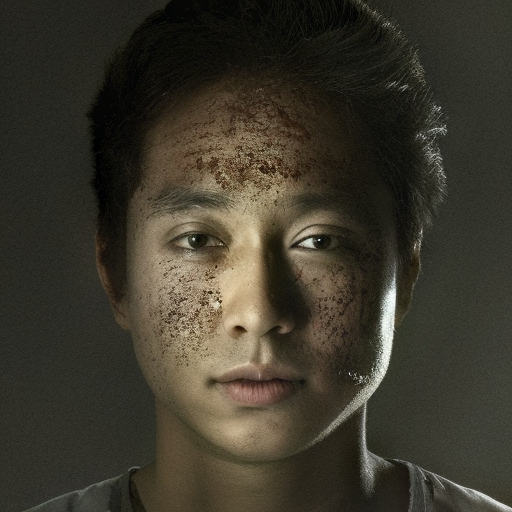}~
		&\includegraphics[width=0.115\textwidth]{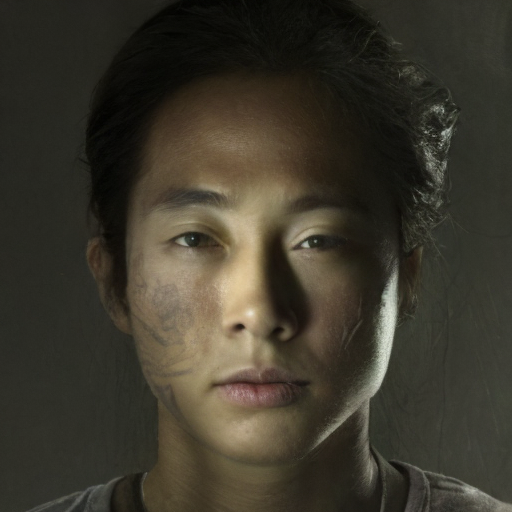}~
    &\includegraphics[width=0.115\textwidth]{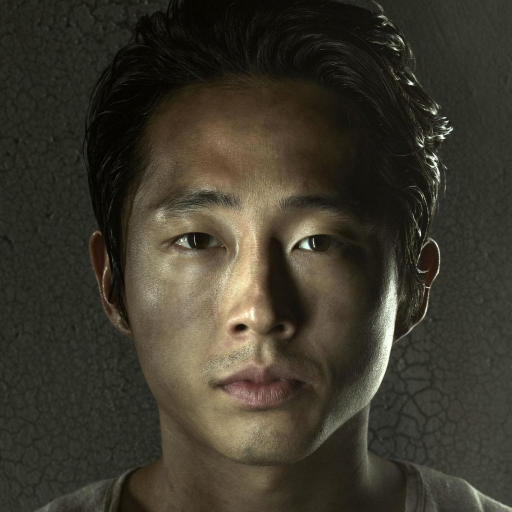}\\
    &\includegraphics[width=0.115\textwidth]{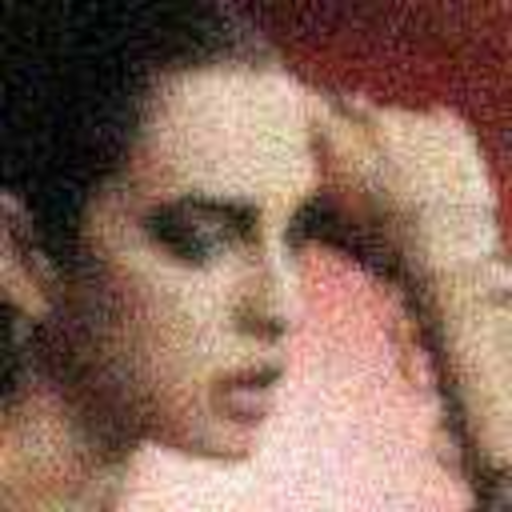}~
    &\includegraphics[width=0.115\textwidth]{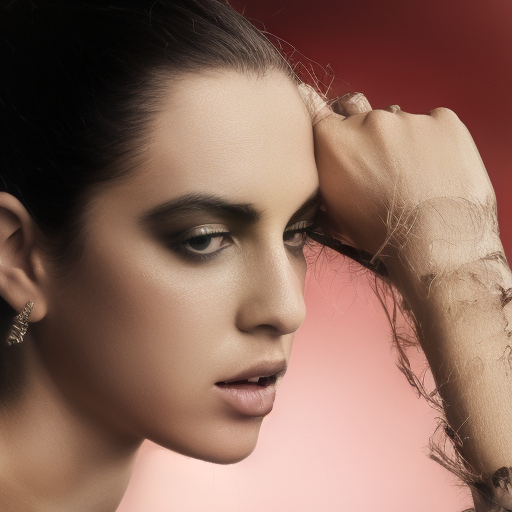}~
		&\includegraphics[width=0.115\textwidth]{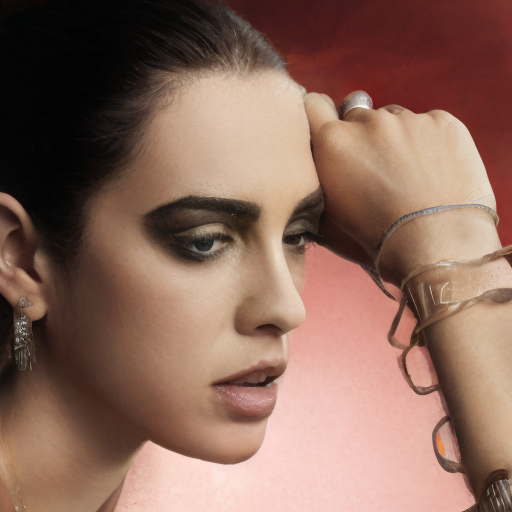}~
    &\includegraphics[width=0.115\textwidth]{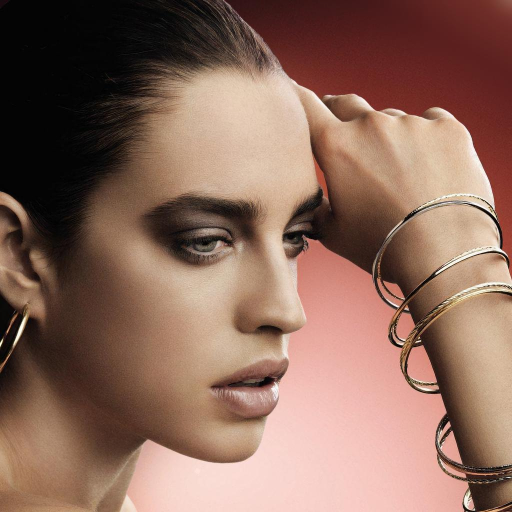}\\
&(a) Input &(b) DiffBIR \cite{lin2023diffbir} & (c) Ours & (d) Ground-Truth\\
	\end{tabular}
	\caption{Comparison of 4× blind super-resolution (SR) using DiffBIR \cite{lin2023diffbir} (b) and our proposed approach (c).}
 \vspace{-0.5cm}
	\label{fig:fig1}
\end{figure}

In recent years, diffusion models \cite{diffusionsurvey} have garnered significant attention for producing visually realistic and diverse image samples. By iteratively denoising data across multiple time steps using a learned denoiser $\epsilon_{\theta}(\cdot)$ (or score estimator), these models establish a powerful prior for generative tasks. Among various diffusion-based methods, Latent Diffusion Models (LDMs) \cite{rombach2022high}, which perform diffusion in a compressed latent representation, have become particularly popular due to their computational scalability and efficiency. Over the past few years, diffusion-based approaches have made rapid progress in inverse problems such as super-resolution \cite{SR3,SR3+,li2022srdiff,cdpmsr,resdiff,sde-sr,diffir,difface,wang2023dr2,idm,ilvr,snips,DDRM,dps,blinddps,ccdf,pgdm,gdp,red-diff,osediff,lin2023diffbir,wang2024exploiting,wu2024seesr,promptfix,DDNM}, deblurring \cite{PnPSGS,deblurdpm,indi,diracdiffusion,adir,feng2023score,bird,dmplug}, JPEG restoration \cite{saharia2022palette,DriftRec,jpeg-ddrm}, low-light enhancement \cite{AnlightenDiff,lowlight}, and multi-modal image fusion \cite{Dif-Fusion,CrossDiff,VDMUFusion}. In particular, building on LDMs, existing methods for inverse problems can be broadly categorized into two groups. The first category \cite{wang2024exploiting,lin2023diffbir,wu2024seesr,promptfix,osediff} fine-tunes a pretrained LDM for a specific restoration task (e.g., image super-resolution), typically yielding a modified denoiser
$\epsilon_{\theta_{IR}}(\cdot)$
that takes the degraded measurement $\bm{y}$ as an additional input.
By adapting the model parameters to specific degradations, these methods often achieve strong task-focused performance and improved controllability, but this specialization may limit their flexibility for rapidly changing or diverse degradation types.
In contrast, the second category \cite{psld,p2l,mpgd,resample,treg,ldir,stsl,silo} keeps the pretrained LDM unchanged, focusing on altering only the inference stage to incorporate measurement constraints.
By preserving the generative capacity of the original LDM and maintaining data fidelity, this approach offers greater adaptability for a variety of inverse problems.

Despite the impressive properties of the second category, several limitations remain.
Firstly, most of them rely on a pre-defined degradation operator. Consequently, they struggle with complex or unknown degradations that do not fit into standard analytical models.
Moreover, providing reliable data-consistency guidance in LDM is particularly difficult as some methods impose strong guidance and may end-up smoothing the high-frequency textures generated by LDMs, whereas others may compromise data fidelity.
Finally, most of them require decoding latent representations back to the pixel domain for guidance at every sampling iteration, which introduces high computational and memory overhead and consequently slows the inference process.

To address these limitations, we draw inspiration from the fundamental properties of the wavelet transform, which provide a strictly invertible decomposition into coarse and detail components. By replacing the fixed filtering operations in the wavelet transform with learnable neural modules, we develop an invertible neural network (INN), whose forward transform simulates the degradation process, and inverse transform achieves reconstruction.
In principle, this forward transform factorizes the ground-truth image into a coarse component and corresponding details, representing the measurements and lost details, respectively.  Due to its perfect reconstruction property, the inverse transform can fully recover the original ground-truth image.
However, in practical inverse problems, only the degraded measurements are available. To compensate for the missing details, we incorporate detail information sampled from the LDM. Consequently, the reconstruction merges the fidelity of the measured data with the fine structures contributed by the LDM sample.

To incorporate this proposed invertible design into the LDM sampling pipeline, we propose an INN-Guided Probabilistic Diffusion Algorithm for Latent diffusion models (LatentINDIGO), which comprises two approaches.\footnote{The project code will be made publicly available upon acceptance.} The first approach, LatentINDIGO-PixelINN, applies a wavelet-inspired invertible neural network in the pixel domain (PixelINN). At each iteration, we decode the current estimated clean latent representation into an image, decompose it into a degraded component and a detail component, and then fuse the detail component with the actual measurements $\bm{y}$ to produce a PixelINN-optimized result. Then the sampling is guided via gradient updates to ensure consistency and high-quality details.
The second approach, LatentINDIGO-LatentINN, places the invertible network (LatentINN) entirely in the latent space, modeling the relationship between encoded representations of original and degraded images. This design eliminates the need to decode latent representations back to image pixels at each iteration, thereby avoiding frequent conversions. In this way, we significantly reduce computational and memory overhead.  Both methods enjoy a more stable consistency step than other methods due to the use of an invertible architecture with its joint forward and inverse constraints.
Furthermore, both proposed approaches incorporate two strategies to accommodate unknown degradations and maintain perceptual quality. First, we alternate between updating intermediate latent variables under the guidance of our INN and refining the forward model of the INN to handle unknown degradations, enabling the framework to adapt more flexibly to real-world unknown degradations.
Second, to keep the intermediate latent variables on the natural image manifold after each INN-guidance update, we incorporate a regularization step that mitigates potential distribution shifts, thereby ensuring high-quality reconstructions.

We summarize our contributions as follows:

\begin{itemize} \item  Building on the principle of wavelet-inspired invertibility, we propose two approaches, LatentINDIGO-PixelINN and LatentINDIGO-LatentINN, which integrate INNs into LDM frameworks operating in the pixel and latent domains, respectively. Both approaches ensure data consistency while preserving rich details during the LDM sampling, and do not rely on the knowledge of an explicit analytical form of the degradation operator. This makes our methods suitable for complex real-world scenarios.

\item Our on-the-fly refinement mechanism enables the framework to adapt dynamically to arbitrary degradations during testing, thereby improving real-world applicability. Also, we introduce a regularization step to counterbalance the shifts caused by the INN-guidance update, ensuring that the intermediate latent variables remain near the natural image manifold.

\item Experiments (see Fig. \ref{fig:fig1}) demonstrate that our algorithms yield competitive results compared with leading methods, both quantitatively and visually, on synthetic and real-world low-quality images, and can be adapted to arbitrary output sizes.

\item Our proposed off-the-shelf algorithm can be easily integrated into existing latent diffusion pipelines to enhance IR performance without requiring additional retraining or fine-tuning of LDMs, thereby demonstrating the effectiveness and flexibility of our framework.
\end{itemize}

\section{Related work}
\begin{figure*}[t]
    \centering
        \subfloat[]{
        \includegraphics[width=0.42\textwidth]{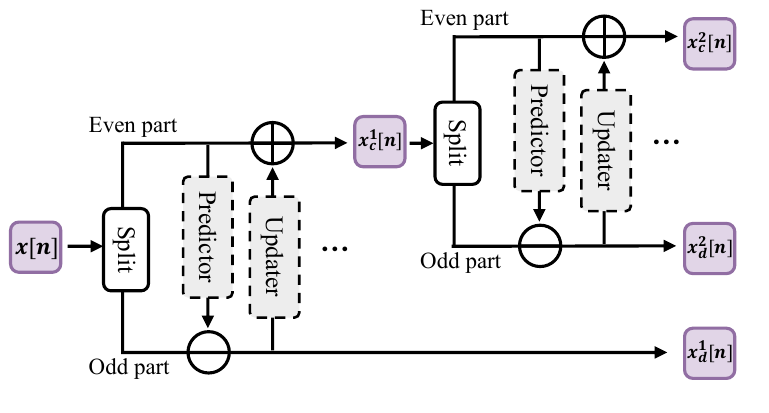}
        \label{fig:details_a}

    }
        \subfloat[]{
        \includegraphics[width=0.42\textwidth]{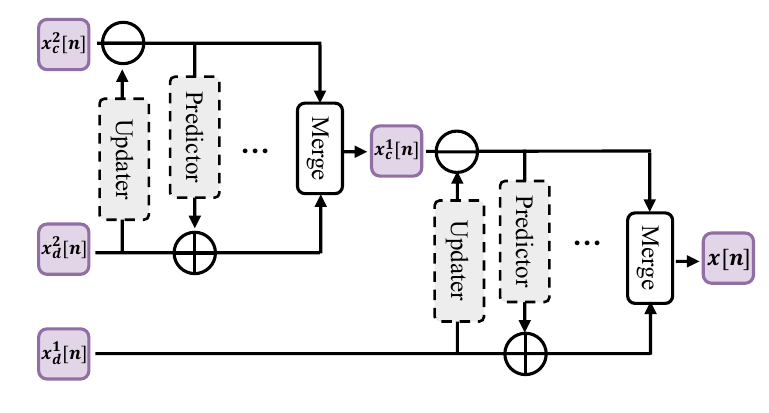}}
        \vspace{-0.2cm}
    \caption{Illustration of two-level lifting scheme on a one-dimensional signal $\bm{x} \textlbrackdbl n \textrbrackdbl$. (a) shows the forward transform: at each level, the input is split into odd and even samples, and predict/update operations are applied to generate coarse and detail components. The coarse component is then passed to the next level for further decomposition, forming a multi-resolution representation. (b) shows the inverse transform, which reverses the process by using the same predict/update operations and a merge operator to combine the coarse and detail components at each level, ultimately reconstructing the original signal. In a wavelet-inspired invertible neural network, the predict and update operators are implemented using trainable neural networks. }
    \label{fig:lifting}
    \vspace{-0.2cm}
\end{figure*}

\subsection{Diffusion Models}
Diffusion models \cite{diffusionsurvey} progressively corrupt data using a pre-defined noise schedule, and learn a denoising objective (or score function) to reverse this corruption, thereby forming a generative model. Both the forward corruption and reverse generative processes can be rigorously formulated within the continuous-time framework of Itô stochastic differential equations (SDEs). In practice, these continuous-time formulations are often discretized into tractable steps, leading to algorithms such as Denoising Diffusion Probabilistic Models (DDPM) \cite{ddpm}.
In DDPM, given a clean image $\bm{x}_0$ and a variance schedule $\{\beta_t\}_{t=1}^T$, the forward diffusion process is defined as:
\begin{equation}
q(\bm{x}_t \mid \bm{x}_{t-1}) = \mathcal{N}(\bm{x}_t; \sqrt{1-\beta_t}\,\bm{x}_{t-1}, \beta_t \mathbf{I}),
\end{equation}
resulting in $\bm{x}_T$ converging to an isotropic Gaussian for sufficiently large \(T\). During training, a neural network $\epsilon_\theta$ is trained to predict the noise $\epsilon_{train}$ added at step $t$ by minimizing $
\mathbb{E}_{t,\bm{x}_0,\epsilon_{train}}[\|\epsilon_{train} - \epsilon_\theta(\bm{x}_t,t)\|_2^2].
$
At inference, DDPM samples iteratively from Gaussian noise $\bm{x}_T \sim \mathcal{N}(0,I)$, using the reverse diffusion step:
\begin{equation}
\bm{x}_{t-1} = \frac{1}{\sqrt{\alpha_t}}\left(\bm{x}_t - \frac{1-\alpha_t}{\sqrt{1-\bar{\alpha}_t}}\epsilon_{\theta}(\bm{x}_t,t)\right) + \sigma_t \bm{\epsilon},
\label{ddpm}
\end{equation}
where $\alpha_t = 1 - \beta_t$, \(\bar{\alpha}_t = \prod_{i=1}^t \alpha_i\), $\sigma_t= \sqrt{\frac{\,1 - \bar{\alpha}_{t-1}\,}{\,1 - \bar{\alpha}_t\,} \;\beta_t}$ and \(\bm{\epsilon} \sim \mathcal{N}(0,I)\). Often the reverse diffusion step is computed as follows:
\begin{equation}
\bm{x}_{t-1}  = \frac{\sqrt{\alpha_{t-1}} \beta_t}{1-\bar{\alpha}_t} {\bm{x}}_{0,t} + \frac{\sqrt{\alpha_t} \left( 1 - \bar{\alpha}_{t-1} \right)}{1 - \bar{\alpha}_t} \bm{x}_t+ {\sigma}_t \epsilon
\label{ddpm_1}
\vspace{-0.1cm}
\end{equation}
where
\begin{equation}
\bm{x}_{0,t}  = \frac{1}{\sqrt{\bar\alpha_t}}(\bm{x}_{t} - \sqrt{1 - \bar\alpha_t}
\bm{\epsilon}_\theta(\bm{x}_t, t) )
\label{ddpm_2}
\vspace{-0.1cm}
\end{equation}
is the estimated clean image given $\bm{x}_{t}$.

To address computational scalability challenges faced by pixel-space models, Latent Diffusion Models (LDMs) \cite{rombach2022high} perform diffusion in a compressed latent representation. Specifically, an image \( \bm{x} \) is first encoded into a latent representation \( \bm{z} = \mathcal{E}(\bm{x}) \). A diffusion model is then trained in this low-dimensional latent space. During sampling, the reverse diffusion process generates latent samples \( \bm{z}_0 \), which are subsequently decoded back to the original image space through a decoder: \( \bm{x}_0 = \mathcal{D}(\bm{z}_0) \).

\vspace{-0.2cm}
\subsection{Solving IR with (Latent) Diffusion Models}
\label{related_work_ir_ldm}
Diffusion models have been increasingly adopted for solving inverse problems in imaging. Generally, approaches leveraging diffusion models for inverse problems can be categorized into two main groups.

One line of work \cite{SR3,SR3+,li2022srdiff,cdpmsr,resdiff,sde-sr,resshift}
emphasizes retraining the denoiser (or score estimator) of diffusion models to explicitly incorporate characteristics and constraints specific to the inverse problem considered.
These specialized diffusion models often achieve remarkable reconstruction quality but demand additional computational overhead and retraining complexity.
With the emergence of LDMs, several recent methods \cite{osediff,lin2023diffbir,wang2024exploiting,wu2024seesr,promptfix,pasd,sinsr,supir} have proposed fine-tuning LDMs specifically for image super-resolution tasks, demonstrating enhanced reconstruction quality and improved controllability.
Specifically, they fine-tune the original LDM denoiser \(\epsilon_{\theta}(\bm{z}_t, t, \bm{\gamma}_{\text{text}})\) to obtain a version \(\epsilon_{\theta_{\text{IR}}}(\bm{z}_t, t, \bm{\gamma}_{\text{text}}, \bm{y})\) that is specifically adapted for image reconstruction given the measurements \(\bm{y}\).
The conditioned reconstruction result $\bm{x}_0$ from the observed low-resolution image $\bm{y}$ can be obtained via the following iterative sampling process:
\begin{equation}
    \bm{z}_{t-1} = \frac{1}{\sqrt{\alpha_t}}\left(\bm{z}_t - \frac{1 - \alpha_t}{\sqrt{1 - \bar{\alpha}_t}}\epsilon_{\theta_{IR}}(\bm{z}_t, t, \bm{\gamma}_{text}, \bm{y})\right) + \sigma_t \bm{\epsilon},
\end{equation}
where $\bm{\epsilon}$ is samped from $\mathcal{N}(0,\bm{I})$,
and $\epsilon_{\theta_{IR}}(\cdot)$ denotes the fine-tuned denoiser specifically adapted for image reconstruction.

An alternative research direction \cite{ilvr,ccdf,snips,DDRM,jpeg-ddrm,DDNM,yang2023pgdiff,gdp,red-diff,pgdm,dps,blinddps,dpps,Laroche_2024_WACV,FPS-SMC,Xu2025RethinkingDP,indigoplus} preserves the pretrained unconditional diffusion model and modifies only the inference procedure, enabling conditional sampling from the posterior given observed measurements. In this scheme, the pretrained diffusion model provides a generative prior, while data consistency is enforced during the sampling process.
From a score-based sampling perspective, we can further express the conditional distribution as:
\begin{equation}
    \nabla_{\bm{x}_t}\log p(\bm{x}_t|\bm{y}) = \nabla_{\bm{x}_t}\log p(\bm{x}_t) + \nabla_{\bm{x}_t}\log p(\bm{y}|\bm{x}_t),
    \label{eq:condition_score_theory}
\end{equation}
where the first term $\nabla_{\bm{x}_t}\log p(\bm{x}_t)$ can be estimated using the pre-trained denoiser $\epsilon_{\theta}(\cdot)$ of the diffusion model (or equivalently a score estimator)
and the second term $\nabla_{\bm{x}_t}\log p(\bm{y}|\bm{x}_t)$ is intractable for IRs and is therefore approximated. Various strategies have been proposed to approximate this term. For instance, DPS \cite{dps} adds a gradient-based data-consistency correction after each unconditional DDPM update:
\begin{equation}
    \bm{x}_{t-1} = \hat{\bm{x}}_{t-1} - \rho\nabla_{\bm{x}_t}\|\bm{y}-\mathcal{H}({\bm{x}}_{0,t})\|_2^2,
\end{equation}
where $\hat{\bm{x}}_{t-1}$ represents an intermediate state obtained via standard unconditional DDPM sampling as in Eq. \ref{ddpm}. Moreover, ${\bm{x}}_{0,t}$ represents a direct estimation of the clean data from the noisy sample ${\bm{x}}_{t}$, derived from Tweedie’s formula (Eq. \ref{ddpm_2}), and $\rho$ controls the step size of data consistency enforcement.

To solve IR problems with LDMs, recent studies \cite{psld,p2l,mpgd,resample,treg,ldir,stsl} extend the idea of DPS \cite{dps} from the image domain into the latent domain. Their baseline approach, termed LDPS \cite{psld}, straightforwardly adapts DPS by applying the data-fidelity correction in latent space as follows:
\begin{align}
{\bm{z}_{t-1}}= \hat{\bm{z}}_{t-1}-\lambda \nabla_{\bm{z}_{t}}\|\bm{y} - \mathcal{H}(\mathcal{D}(\bm{z}_{0,t})\|_2^2,
\label{ldps}
\end{align}
where $\lambda$ is the step size, $\hat{\bm{z}}_{t-1}$ represents an intermediate state obtained via standard DDPM sampling and $\mathcal{D}$ is the decoder mapping latents back to the image domain. Building on LDPS, subsequent works \cite{psld,p2l,mpgd,resample,treg,ldir,stsl} explore stronger regularization, prompt optimization, and resampling strategies, leading to higher-fidelity and more stable reconstructions.

% \vspace{-0.1cm}
\section{LatentINDIGO}

\begin{figure*}[t]
    \centering
    \subfloat[Forward transform of PixelINN.]{
        \includegraphics[width=0.48\textwidth]{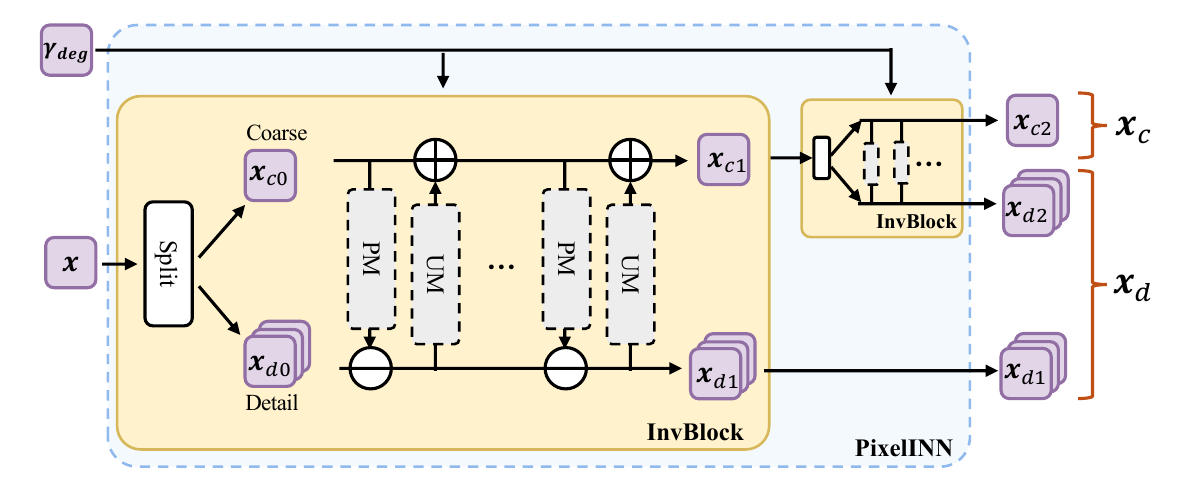}
        \label{fig:details_a}
    }
        \subfloat[Inverse transform of PixelINN.]{
        \includegraphics[width=0.48\textwidth]{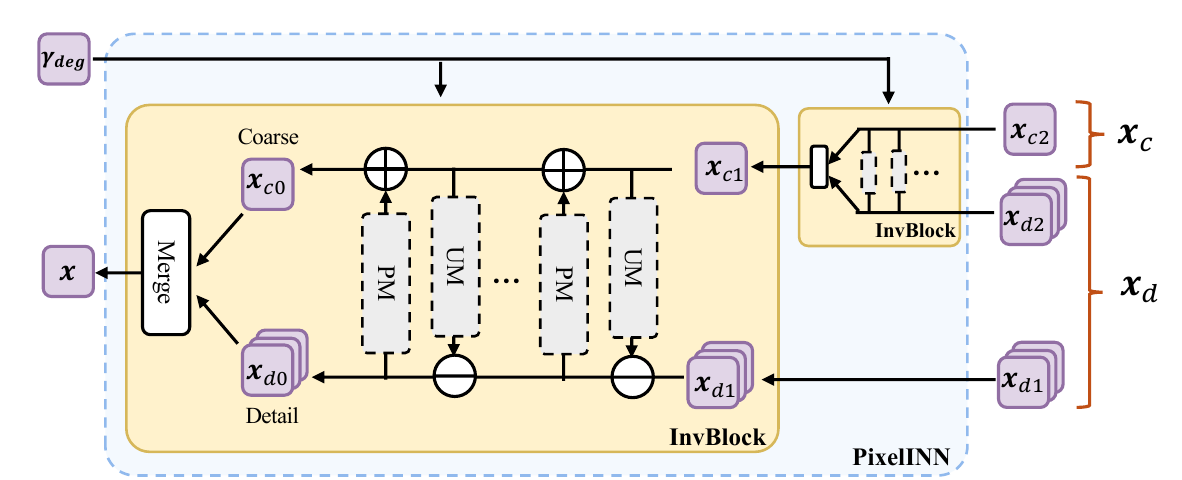}
        \label{fig:details_a}
    }
    \caption{Architecture of the proposed PixelINN. (a) Forward pass: The first lifting level splits the input image \(\bm{x}\) into coarse \(\bm{x}_{c0}\) and detail \(\bm{x}_{d0}\) components. The predict–update modules (PM and UM) then refine these components, producing \(\bm{x}_{c1}\) and \(\bm{x}_{d1}\). The next lifting level processes \(\bm{x}_{c1}\) similarly, yielding a new coarse–detail pair \(\bm{x}_{c2}, \bm{x}_{d2}\). (b) Inverse pass: The same PM and UM modules, combined with a merge operator, are used to recombine the coarse and detail subbands level by level. Starting from the final-stage subbands, the network reverses each lifting level and, owing to its perfect reconstruction property, can ultimately recover the original image \(\bm{x}\). PixelINN is conditioned to the vector $\bm{\gamma}_{\mathrm{deg}}$ which embeds information about the degradation model.}
    \label{fig:pinn}
    \vspace{-0.3cm}
\end{figure*}

\subsection{Overview}
We propose two approaches to address the IR problem using LDM: LatentINDIGO-PixelINN and LatentINDIGO-LatentINN. At the heart of both methods is the use of wavelet-inspired invertible neural networks (INNs) to integrate the LDM-generated samples with the measurements. These networks are perfectly invertible because their construction is based on the lifting scheme, which is also used to construct the wavelet transform (WT) \cite{daubechies1998factoring,winnet}. The lifting scheme is characterized by an invertible `split' operator followed by predict and update operations. Exact invertibility is achieved by inverting the order of operations and the signs as shown in Fig. \ref{fig:lifting} and this invertibility is guaranteed independently on how the predict and update operations are implemented. As in the case of the WT, the scheme splits the signal in coarse and detail components and the decomposition can be repeated on the coarse version. In a wavelet-inspired INN, the predict and update operations are implemented using trainable neural networks. In our work, we train the forward part of the INN so that the coarse component mimics the degradation of the original image while the corresponding details represent the information lost in the degradation process. The flexibility of the INN enables us to model a broad range of complex degradations. Moreover, in our work, we propose that the details are enriched by the LDM samples, thereby merging the fidelity of measured data with the fine structures contributed by the LDM through the inverse INN.
LatentINDIGO-PixelINN, applies the invertible transformation directly in the pixel domain, leveraging a lifting-inspired network to model various image degradations.
LatentINDIGO-LatentINN, instead, places the invertible network within the latent space of the LDM, modeling the relationship between encoded representations of original and degraded images. Here, images are initially mapped into a compressed representation via the VAE encoder, and the INN operations are subsequently performed on these latent features, thereby avoiding frequent conversions.

\subsection{LatentINDIGO-PixelINN}
In the following, we present the design of our PixelINN, highlighting how its invertibility provides a framework for analyzing forward degradations and inverse reconstructions in inverse problems.
\subsubsection{The Architecture of PixelINN}
Fig.~\ref{fig:pinn} illustrates the proposed PixelINN framework. In particular, Fig.~\ref{fig:pinn}(a) depicts the {forward} pass, where the input image \(\bm{x}\) is successively split into coarse and detail parts, while Fig.~\ref{fig:pinn}(b) shows the {inverse} pass, which recombines these parts back into a complete reconstruction. Both forward and inverse processes share the same network parameters, ensuring strict invertibility. By leveraging this invertibility, we interpret the forward transform \( g_{\bm{\Theta}}(\cdot) \) as a simulator of the degradation process, while the inverse transform \( g_{\bm{\Theta}}^{-1}(\cdot) \) serves as the reconstruction operator, allowing the model to effectively learn and invert complex degradations.

\textbf{Multi-Level Lifting Structure}. Similar to classical wavelet transforms, PixelINN can perform multiple levels of decomposition to capture multi-scale information. In Fig.~\ref{fig:pinn}(a), each Invblock corresponds to one level of lifting-based decomposition, consisting of a split operator and a sequence of predict module (PM) and update module (UM) pairs. The coarse output of one level (e.g., \(\bm{x}_{c1}\)) can be further split and processed by the next level.

As shown in Fig.~\ref{fig:pinn}(a), the first lifting level splits the input image $\bm{x}$ into a coarse component \(\bm{x}_{c0}\) and detail component \(\bm{x}_{d0}\). We implement the split operator with either a redundant or a non-redundant wavelet transform. Subsequently, the predict–update modules (PM and UM) refine these coarse and detail parts, producing updated components \(\bm{x}_{c1}\) and \(\bm{x}_{d1}\). The next lifting level then processes \(\bm{x}_{c1}\) similarly, yielding a new coarse-detail pair \(\bm{x}_{c2}, \bm{x}_{d2}\). The inverse pass, shown in Fig.~\ref{fig:pinn}(b), employs the same PM and UM modules together with a merge operator, implemented by the inverse wavelet transform.

\begin{figure}[t]
    \centering
    \includegraphics[width=0.45\textwidth]{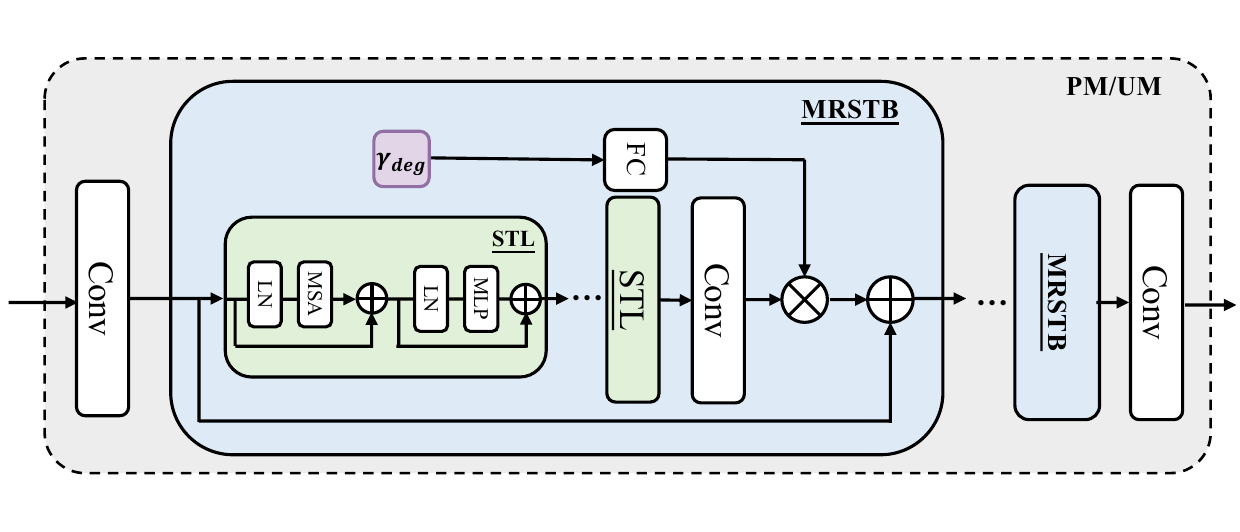}
    \vspace{-10pt}
\caption{Architecture of the proposed PM/UM. Each PM/UM starts with a convolutional layer for channel adjustment, followed by our Modulated Residual Swin Transformer Blocks (MRSTBs), and ends with another convolutional layer that refines the feature maps and projects them into the desired output channels. The MRSTB module is composed of Swin Transformer Layers (STLs) with multi-head self-attention (MSA) and multi-layer perceptron (MLP), along with a learnable modulation mechanism conditioned on the degradation embedding \(\bm{\gamma}_{\mathrm{deg}}\).}
    \label{fig:pmum}
    \vspace{-0.2cm}
\end{figure}

\textbf{The Architecture of PM and UM}.
Fig.~\ref{fig:pmum} illustrates our proposed architecture of predict and update modules, which, different from other INN architecures \cite{revnet,winnet}, integrates convolutional operations with Swin Transformer layers (STLs) \cite{liang2021swinir}, guided by the degradation embedding $\bm{\gamma}_{\mathrm{deg}}$. Each PM/UM begins with a convolutional layer that adjusts channel dimensions, followed by our proposed Modulated Residual Swin Transformer Blocks (MRSTBs), and concludes with another convolutional layer to refine the output before passing it to the next module.
As shown in the blue region, each MRSTB consists of two key components: multiple STLs for long-range dependency modeling, and a learnable modulation mechanism conditioned on the degradation embedding $\bm{\gamma}_{\mathrm{deg}}$.
Instead of assuming a predefined degradation process, our approach estimates $\bm{\gamma}_{\mathrm{deg}}$ using a pre-trained degradation estimation module $\mathcal{F}(\cdot)$, formulated as: $\bm{\gamma}_{\mathrm{deg}} = \mathcal{F}(\bm{y})$.
The modulation applies a channel-wise scaling to the feature map, with the scaling factors computed from $\bm{\gamma}_{\mathrm{deg}}$ via a fully connected (FC) layer. This adaptive modulation enables the network to dynamically emphasize or suppress specific feature channels in response to different degradation types and levels.

\subsubsection{Training Strategy for PixelINN}

Given a clean image \(\bm{x}\) and its corresponding degraded observation \(\bm{y}\), we train the network to approximate the degradation through the forward transform \(g_{\mathbf{\Theta}_{pix}}(\cdot)\), while reconstructing the original clean image from degraded measurements via the inverse transform \(g_{\mathbf{\Theta}_{pix}}^{-1}(\cdot)\).
Specifically, the forward process decomposes the clean image \(\bm{x}\) into coarse and detail components:
\begin{equation}
[\bm{x}_c, \bm{x}_d] = g_{\mathbf{\Theta}_{pix}}(\bm{x}, \bm{\gamma}_{\mathrm{deg}}).
\end{equation}
Subsequently, we replace the coarse component $\bm{x}_c$ with the degraded observation \(\bm{y}\), and apply the inverse transform:
\begin{equation}
{\bm{x}_{inv}} = g_{\bm{\Theta}_{pix}}^{-1}(\bm{y}, \bm{x}_d, \bm{\gamma}_{\mathrm{deg}}).
\end{equation}
To ensure that the coarse component $\bm{x}_c$ aligns with the degraded image \(\bm{y}\), we introduce a supervised forward loss:
\begin{align}
\begin{split}
\label{eq:lossfunc}
\vspace{-0.1cm}
L_{\text{forw}}\bigl(\mathbf{\Theta}_{pix}\bigr)
&= \frac{1}{N} \sum_{i=1}^{N} \left\|\bm{x}_c^{(i)} - \bm{y}^{(i)}\right\|_2^2
\vspace{-0.1cm}
\end{split}
\end{align}
where \(N\) denotes the total number of training samples.
In addition, to improve the reconstruction quality and reduce potential artifacts in the inverse process, we introduce a loss on the reconstructed image:

\begin{align}
\begin{split}
\label{eq:lossfunc}
\vspace{-0.1cm}
L_{inv}\left ( {\mathbf{\Theta}_{pix}}  \right )=\frac{1}{N}\sum_{i=1}^{N}\left \|\bm{x}_{inv}^{(i)}-\bm{x}^{ (i) } \right \|_{2}^{2}.
\vspace{-0.1cm}
\end{split}
\end{align}

Therefore, the total loss for PixelINN training is defined as a combination of forward and inverse objectives with a weighting factor $\lambda_{inv}$ :
\begin{equation}
\mathcal{L}_{\text{PixelINN}} = \mathcal{L}_{\text{forw}} + \lambda_{inv} \mathcal{L}_{\text{inv}}.
\end{equation}
Thus, this strategy ensures that the model accurately simulates realistic degradation processes while effectively yielding high-quality reconstruction results.
\begin{algorithm}[]
           \caption{LatentINDIGO-PixelINN}
           \label{alg:indigo_latent}
            \begin{algorithmic}[1]
             \Require Corrupted image $\bm{y}$,
pretrained PixelINN $ g_{\mathbf{\Theta}_{pix}}(\cdot)$, estimated degradation embedding $\bm{\gamma}_{deg}$.
             \State $\bm{z}_T \sim \mathcal{N}(\bzero, \bI)$
              \For{$t=T$ {\bfseries to} $1$}
\State {{$\bm{z}_{0,t}  = \frac{1}{\sqrt{\bar\alpha_t}}(\bm{z}_{t} - \sqrt{1 - \bar\alpha_t}
\bm{\epsilon}_\theta(\bm{z}_t, t, \cdot) )$}}
 \State      {\color{black}{$[\bm{x}_{c,t};\bm{x}_{d,t}] =
 g_{\mathbf{\Theta}_{pix}}(\mathcal{D}(\bm{z}_{0,t}),\bm{\gamma}_{deg})$}}
\State       {\color{black}{${\bm{x}}_{inv,t}=
g^{-1}_{\mathbf{\Theta}_{pix}}(\bm{y},\bm{x}_{d,t},\bm{\gamma}_{deg})$}}
\State        {\color{black}{$\ell_{\mathrm{forw}}=\|{\bm{x}_{c,t}- \bm{y}}\|_2^2 $}}
\State        {\color{black}{$\ell_{\mathrm{inv}}=
 \|{\varphi({\bm{x}}_{inv,t})- \varphi(\mathcal{D}(\bm{z}_{0,t}))}\|_2^2$}}
\State        {\color{blue}{$\tilde{\bm{z}}_{0,t}=\bm{z}_{0,t} - { {\zeta}}\nabla_{\bm{z}_{0,t}}(\alpha_{forw}\ell_{\mathrm{forw}}+\alpha_{inv}\ell_{\mathrm{inv}}) $}} \Comment{{\textcolor{gray}{PixelINN Guidance}}}

\State        {\color{blue}{$\hat{\bm{z}}_{0,t} = \mathcal{E}(\mathcal{D}(\tilde{\bm{z}}_{0,t}))$}} \Comment{{\textcolor{gray}{Regularization}}}
\State  $\bm{z}_{t-1}  = \frac{\sqrt{\alpha_{t-1}} \beta_t}{1-\bar{\alpha}_t} \hat{\bm{z}}_{0,t} + \frac{\sqrt{\alpha_t} \left( 1 - \bar{\alpha}_{t-1} \right)}{1 - \bar{\alpha}_t} \bm{z}_t+ {\sigma}_t \epsilon  $
          \State        {\color{blue}{$\mathbf{\Theta}_{pix}=\mathbf{\Theta}_{pix} - l \nabla_{\mathbf{\Theta}_{pix}} \|{\bm{x}_{c,t}- \bm{y}}\|_2^2$}} \Comment{{\textcolor{gray}{Refinement}}}
              \EndFor
              \State \textbf{return} $\mathcal{D}(\bm{z}_{0})$
            \end{algorithmic}
    \end{algorithm}

\vspace{-0.1cm}
\subsubsection{Inference Process of LatentINDIGO-PixelINN}
Our LatentINDIGO-PixelINN framework is detailed in Algorithm~\ref{alg:indigo_latent}.
We denote the learned LDM denoiser by \(\bm{\epsilon}_\theta(\bm{z}_t, t, \cdot)\), where `\(\cdot\)' indicates our method’s capacity to accommodate various pretrained or fine-tuned denoisers of LDMs. In the simplest scenario, \(\bm{\epsilon}_\theta\) depends solely on \((\bm{z}_t, t)\) or additionally takes text prompts via $\epsilon_{\theta}(\bm{z}_t, t, \bm{\gamma}_{text})$. More specialized, finetuned versions, as discussed in Section \ref{related_work_ir_ldm}, may also incorporate measurement inputs \(\epsilon_{\theta_{IR}}(\bm{z}_t, t, \bm{y})\) and corresponding text prompts \(\epsilon_{\theta_{IR}}(\bm{z}_t, t, \bm{y}, \bm{\gamma}_{text})\), thereby adapting the original LDMs to inverse problems. Notably, these pretrained LDMs can be integrated into our framework to potentially improve data fidelity and visual quality for diverse (including unseen) degradations, all without requiring retraining or finetuning of the learned denoiser.

At each step during sampling, we first estimate the noise present in the current noisy latent variable $\bm{z}_t$ with \(\bm{\epsilon}_\theta(\bm{z}_t, t, \cdot)\) to predict a denoised latent variable $\bm{z}_{0,t}$, serving as an initial reference for subsequent guidance steps. We then decode $\bm{z}_{0,t}$ into its image domain representation $\mathcal{D}(\bm{z}_{0,t})$, which is processed by our pretrained PixelINN to decompose it into two parts: a coarse component $\bm{x}_{c,t}$ approximating degraded measurements, and a detail component $\bm{x}_{d,t}$ containing the high-frequency details lost during degradation, formulated as
\begin{equation}
[\bm{x}_{c,t}, \bm{x}_{d,t}] = g_{\mathbf{\Theta}_{pix}}(\mathcal{D}(\bm{z}_{0,t}),\bm{\gamma}_{\mathrm{deg}}).
\end{equation}
To enforce data fidelity, we replace $\bm{x}_{c,t}$ with observed measurements $\bm{y}$, and reconstruct an INN-enhanced image $\bm{x}_{inv,t}$ via inverse transformation:
\begin{equation}
\bm{x}_{{inv},t}=
g^{-1}_{\mathbf{\Theta}_{pix}}(\bm{y},\bm{x}_{d,t},\bm{\gamma}_{deg})
\end{equation}
which ensures data consistency while effectively recovering lost high-frequency details.
To guide the sampling, we propose a combined loss for updating $\bm{z}_{0,t}$:  an explicit data-consistency constraint in the measurement domain
$\ell_{\mathrm{forw}}=\|\bm{x}_{c,t}-\bm{y}\|_2^2$, which can be seen as an extension of LDPS \cite{dps} with our pretrained PixelINN, and a perceptual loss in the image domain $\ell_{\mathrm{inv}}=\|\varphi(\bm{x}_{inv})-\varphi(D(\bm{z}_{0,t}))\|_2^2$, where $\varphi(\cdot)$ is the pretrained LPIPS \cite{lpips} feature extraction function (see steps 6-8 in Algorithm \ref{alg:indigo_latent}).

\textbf{Regularization.} Although data consistency is maintained in the above steps, the intermediate latent vector $\tilde{\bm{z}}_{0,t}$ may shift off the valid latent manifold.
To address this issue, we re-encode the decoded latent vector as $\hat{\bm{z}}_{0,t} = \mathcal{E}(\mathcal{D}(\tilde{\bm{z}}_{0,t}))$, as a regularization step. We explain this design from two perspectives: Firstly, mapping \(\tilde{\bm{z}}_{0,t}\) back through \(\mathcal{D}\) followed by \(\mathcal{E}\) pushes $\tilde{\bm{z}}_{0,t}$ towards a fixed point of the successive application of the encoding-decoding steps. In fact, we note that the proposed INN $\ell_{\mathrm{inv}}$ guidance combined with this re-encoding strategy can be seen as a variation of the `gluing' mechanism in PSLD \cite{psld}.
Secondly,
because the re-encoding step involves projecting data into a lower-dimensional latent space, it can be viewed as a form of implicit denoising in pixel domain.
Off-manifold artifacts introduced by data-consistency guidance are effectively removed by this re-encoding step, yielding cleaner and more realistic reconstructions.

\textbf{Refinement mechanism.}
Both the pre-trained PixelINN and the degradation estimation module are initially trained on synthetic degradation pairs that may not perfectly align with real-world degradations. To mitigate this discrepancy, we introduce a refinement mechanism during inference. Specifically, at each iteration, given the current estimated clean image, we update the PixelINN parameters to simulate more accurately the observed degradations.
Concretely, we refine the parameters of our PixelINN at inference stage as follows:
\begin{equation}
    \mathbf{\Theta}_{pix}\leftarrow \mathbf{\Theta}_{pix}- l \nabla_{\mathbf{\Theta}_{pix}} \|{\bm{x}_{c,t}- \bm{y}}\|_2^2.
\end{equation}
Through this update (see line 11 in Algorithm \ref{alg:indigo_latent}), the forward and inverse processes, which share the same set of parameters, both benefit from a refined estimation, resulting in more robust guidance for unknown real-world degradations. In our implementation, to maintain computational efficiency, our refinement mechanism is applied only during the first half of the sampling process, i.e., from step $T$ to $T/2$.

\begin{figure*}[t]
    \centering
    \includegraphics[width=0.78\textwidth]{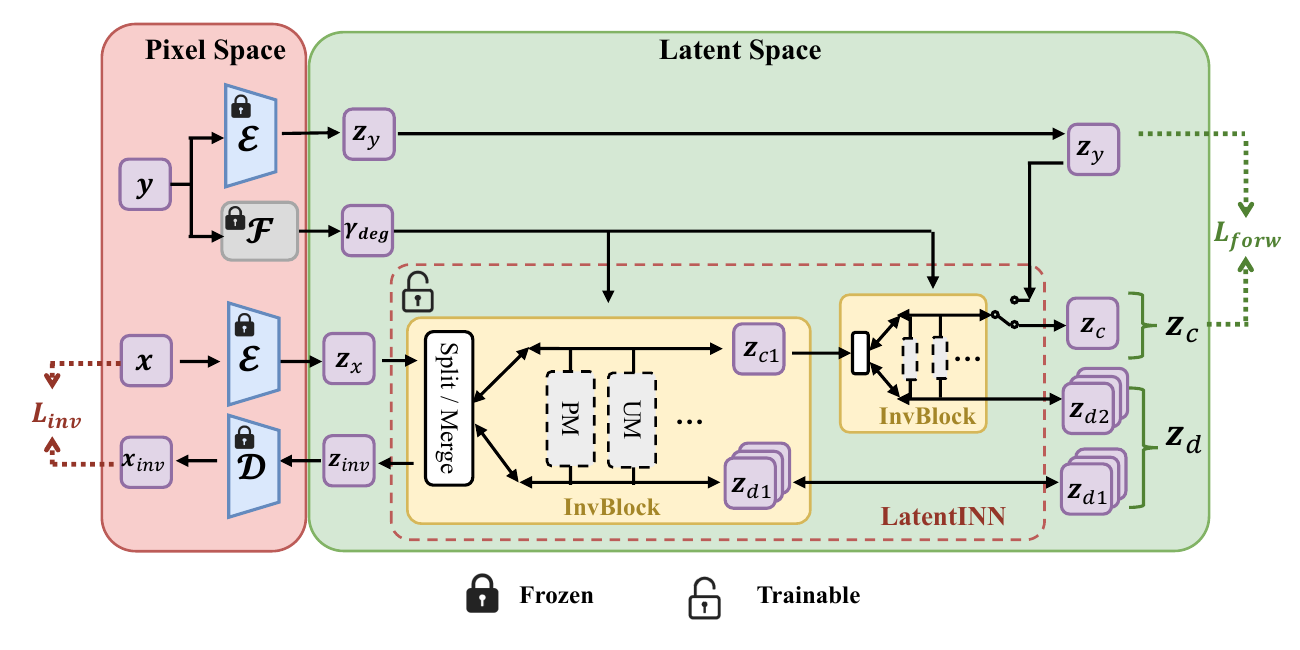}
\vspace{-13pt}
\caption{ The training framework of our LatentINN. During the forward process of our LatentINN, the latent code $\bm{z}_{x}$ of a clean image is transformed into the coarse part $\bm{z}_{c}$ and the detail part $\bm{z}_{d}$. Then we use inverse transform $g_\mathbf{\Theta}^{-1}([\bm{z}_{y};\bm{z}_{d}]) $ with the latent code of the degraded measurement $\bm{z}_{y}=\mathcal{E}(\bm{y})$ to reconstruct  $\bm{z}_{inv}.$}
\label{latent_inn_pipeline}
\vspace{-10pt}
\end{figure*}
\subsection{LatentINDIGO-LatentINN}

While our empirical analyses confirm that our proposed framework significantly outperforms baseline models, we also observe limitations in its memory usage and inference speed.
Specifically, since our PixelINN model estimates image degradation and restoration at the pixel level, the sampling process of LDMs requires converting the current latent code into the pixel domain via the decoder at each iteration, in order to compute the loss and update the latent code. To bypass this issue, we train a latent-level INN so that the entire guidance process remains within the latent space.

\subsubsection{Training LatentINN}
\label{latentinn_training}

The training procedure for LatentINN is illustrated in Fig. \ref{latent_inn_pipeline}. While preserving the architecture of PixelINN, we shift its training from the image domain to the latent domain. Therefore, we first prepare the latent code for clean image $\bm{x}$ and its degraded measurement $\bm{y}$, i.e., $\bm{z}_{{x}}=\mathcal{E}(\bm{x})$ and $\bm{z}_{y}=\mathcal{E}(\bm{y})$\footnote{Here, $\bm{y}$ represents the input after interpolation to match the spatial dimensions of $\bm{x}$.}. In addition, we keep utilizing the pre-trained degradation estimation network $\mathcal{F}(\cdot)$ to model the degradation implicitly with a vector $\bm{\gamma}_{deg}=\mathcal{F}(\bm{y})$. During the forward process of our LatentINN, the input $\bm{z}_x$ is transformed into the coarse part $\bm{z}_{c}$ and the detail part $\bm{z}_{d}$ as follows:
\begin{equation}
[\bm{z}_c, \bm{z}_d] = g_{\mathbf{\Theta}_{lat}}(\bm{z}_x, \bm{\gamma}_{\mathrm{deg}}).
\end{equation}

Subsequently, we use inverse transform with the latent code of the degraded measurement $\bm{z}_{y}=\mathcal{E}(\bm{y})$ to reconstruct  $\bm{z}_{inv}$:
\begin{equation}
{\bm{z}_{inv}} = g_{\mathbf{\Theta}_{lat}}^{-1}(\bm{z}_y, \bm{z}_d, \bm{\gamma}_{\mathrm{deg}}).
\end{equation}

To ensure that the LatentINN network approximates the degradation process in latent space, we train the LatentINN with the following loss function:
\begin{align}
\begin{split}
\label{eq:lossfunc}
\mathcal{L}_{forw}\left ( {\mathbf{\Theta}_{lat}}   \right )=\frac{1}{N}\sum_{i=1}^{N}\left \| \bm{z}_c^{(i)}-\mathcal{E}(\bm{y}^{ (i) }) \right \|_{2}^{2}.
\end{split}
\end{align}
To prevent the latent representation $\bm{z}_{inv}$ from drifting away to a semantically invalid point in latent space, we further introduce an inverse objective to constrain the decoded reconstruction result:
\begin{align}
\begin{split}
\label{eq:lossfunc}
\mathcal{L}_{inv}\left ( {\mathbf{\Theta}_{lat}}  \right )=\frac{1}{N}\sum_{i=1}^{N}\left \|\mathcal{D}(\bm{z}_{inv}^{(i)})-\bm{x}^{ (i) } \right \|_{2}^{2}.
\end{split}
\end{align}
We train our LatentINN by simultaneously utilizing both forward and inverse objectives with a weighting factor $\lambda_{inv}$ :
\begin{equation}
\mathcal{L}_{\text{LatentINN}} = \mathcal{L}_{{forw}} + \lambda_{inv} \mathcal{L}_{{inv}}.
\end{equation}
We will further discuss different training strategies in Section \ref{inn_training}.

\begin{algorithm}[]
           \caption{LatentINDIGO-LatentINN}
           \label{alg:latent}
            \begin{algorithmic}[1]
             \Require Corrupted image $\bm{y}$,
pretrained LatentINN $ g_{\mathbf{\Theta}_{lat}}(\cdot)$, estimated degradation embedding $\bm{\gamma}_{deg}$.
             \State $\bm{z}_T \sim \mathcal{N}(\bzero, \bI)$
\For{$t=T$ {\bfseries to} $1$}
\State {{$\bm{z}_{0,t}  = \frac{1}{\sqrt{\bar\alpha_t}}(\bm{z}_{t} - \sqrt{1 - \bar\alpha_t}
\bm{\epsilon}_\theta(\bm{z}_t, t,  \cdot) )$}}
 \State      $[\bm{z}_{c,t};\bm{z}_{d,t}] =
 g_{\mathbf{\Theta}_{lat}}({\color{black}{ \bm{z}_{0,t}}},\bm{\gamma}_{deg})$
\State       {\color{black}{${\bm{z}}_{inv,t}$}}$=
g_{\mathbf{\Theta}_{lat}}^{-1}({\color{black} \mathcal{E}(\bm{y})},\bm{z}_{d,t},\bm{\gamma}_{deg})$
\State        {\color{blue}$\tilde{\bm{z}}_{0,t}=(1-\alpha) {\bm{z}}_{0,t}+\alpha{{\bm{z}}_{inv,t}} $} \Comment{{\textcolor{gray}{LatentINN Guidance}}}
\State        {\color{blue}{$\hat{\bm{z}}_{0,t} = \mathcal{E}(\mathcal{D}(\tilde{\bm{z}}_{0,t}))$}} \Comment{{\textcolor{gray}{Regularization}}}
\State  $\bm{z}_{t-1}  = \frac{\sqrt{\alpha_{t-1}} \beta_t}{1-\bar{\alpha}_t} \hat{\bm{z}}_{0,t} + \frac{\sqrt{\alpha_t} \left( 1 - \bar{\alpha}_{t-1} \right)}{1 - \bar{\alpha}_t} \mathbf{z}_t+ {\sigma}_t \epsilon  $
          \State        {\color{blue}{$\mathbf{\Theta}_{lat} =\mathbf{\Theta}_{lat} - l \nabla_{\mathbf{\Theta}_{lat} } \|{\bm{z}_{c,t}- \mathcal{E}(\bm{y})}\|_2^2$}} \Comment{{\textcolor{gray}{Refinement}}}
              \EndFor
              \State \textbf{return} $\mathcal{D}(\bm{z}_{0})$
            \end{algorithmic}
    \end{algorithm}
\subsubsection{Inference Process of LatentINDIGO-LatentINN}

Our LatentINDIGO method with LatentINN is shown in Algorithm \ref{alg:latent}. First, as in Algorithm \ref{alg:indigo_latent},
we compute the denoised latent variable
\(\bm{z}_{0,t}\) from \(\bm{z}_t\) to serve as a reference for subsequent guidance.
Next, we decouple $\bm{z}_{0,t}$ into the latent code of coarse and detail part, i.e., $\bm{z}_{c,t}$ and $\bm{z}_{d,t}$, and replace $\bm{z}_{c,t}$ with the latent code $\bm{z}_{y}$ of measurements $\bm{y}$, to maintain data consistency and preserve the details generated by the LDM.
Subsequently, the inverse process of our LatentINN is utilized to transform the combined $\bm{z}_{y}$ and $\bm{z}_{d,t}$ into ${\bm{z}}_{inv,t}$, for further guidance.
Finally, we observe that directly employing an interpolation approach in the latent domain with a scale $\alpha$, is both straightforward and efficient, as illustrated in the sixth line of Algorithm \ref{alg:latent}.
In addition, we preserve the design of regularization update and refinement mechanism proposed in Algorithm \ref{alg:indigo_latent}, to mitigate the deviations caused by the additional modifications to the standard sampling process of LDMs and to enable our LatentINN model to handle more complex degradations in real-world scenarios.

\vspace{-0.2cm}
\subsection{LatentINDIGO for Arbitrary Resolution}

Finally, to support blind image restoration (BIR) at arbitrary resolutions, we integrate our LatentINDIGO framework with patch-based latent diffusion models, as illustrated in Algorithm \ref{alg:patch}. Specifically, lines 3–5 show how each diffusion iteration processes smaller tiles of the latent representation to reduce memory overhead, then merges them via a Gaussian mask into a single complete latent code. Our LatentINN guidance instead operates on the entire latent code at once rather than per patch. The key insight is that our LatentINN is sufficiently lightweight and does not rely on backpropagation-based guidance (e.g., Eq. \ref{ldps} in LDPS \cite{psld} or line 8 in Algorithm \ref{alg:indigo_latent}). Therefore, it can easily operate on the entire latent code. Consequently, our approach ensures consistency with measurements and mitigates the artifacts that could arise from tile-based generation, while preserving the fine details produced by the LDM.
\begin{algorithm}[t]
\caption{LatentINDIGO for Arbitrary Resolution}
\label{alg:patch}
\begin{algorithmic}[1]
\Require
Corrupted image $\bm{y}$,
pretrained LatentINN $ g_{\mathbf{\Theta}_{lat}}(\cdot)$, estimated degradation embedding $\bm{\gamma}_{deg}$,
   Gaussian mask ${mask}$ for each patch.
\State $\bm{z}_T \sim \mathcal{N}(\bzero, \bI)$
\For{$t=T$ {\bfseries to} $1$}
    \For{each patch index $p = 1$ {\bfseries to} $P$}
        \State \(\bm{z}_{0,t}^{({p})} =\tfrac{1}{\sqrt{\bar{\alpha}_t}}
          \bigl(\bm{z}_t^{({p})} - \sqrt{1 - \bar{\alpha}_t}\,\bm{\epsilon}_\theta(\bm{z}_t^{({p})},\,t,\, \cdot )\bigr)\)

    \EndFor

    \State   \(
\bm{z}_{0,t}
=
{\displaystyle \sum_{p=1}^{P}
\Bigl( \bm{z}_{0,t}^{(p)} \;\odot\; \mathrm{mask}^{(p)} \Bigr)}/
{\displaystyle \sum_{p=1}^{P}
\mathrm{mask}^{(p)}}
\)
    \State      $[\bm{z}_{c,t};\bm{z}_{d,t}] =
 g_{\mathbf{\Theta}_{lat}}({\color{black}{ \bm{z}_{0,t}}},\bm{\gamma}_{deg})$
\State       {\color{black}{${\bm{z}}_{inv,t}$}}$=
g_{\mathbf{\Theta}_{lat}}^{-1}({\color{black} \mathcal{E}(\bm{y})},\bm{z}_{d,t},\bm{\gamma}_{deg})$
        \State        {\color{blue}$\tilde{\bm{z}}_{0,t}=(1-\alpha) {\bm{z}}_{0,t}+\alpha{{\bm{z}}_{inv,t}} $} \Comment{{\textcolor{gray}{LatentINN Guidance}}}
\State        {\color{blue}{$\hat{\bm{z}}_{0,t} = \mathcal{E}(\mathcal{D}(\tilde{\bm{z}}_{0,t}))$}} \Comment{{\textcolor{gray}{Regularization}}}
      \State   $\bm{z}_{t-1}  = \frac{\sqrt{\alpha_{t-1}} \beta_t}{1-\bar{\alpha}_t} \tilde{\bm{z}}_{0,t} + \frac{\sqrt{\alpha_t} \left( 1 - \bar{\alpha}_{t-1} \right)}{1 - \bar{\alpha}_t} \mathbf{z}_t+ {\sigma}_t \epsilon  $
            \State        {\color{blue}{$\mathbf{\Theta}_{lat} =\mathbf{\Theta}_{lat} - l \nabla_{\mathbf{\Theta}_{lat} } \|{\bm{z}_{c,t}- \mathcal{E}(\bm{y})}\|_2^2$}} \Comment{{\textcolor{gray}{Refinement}}}
\EndFor
\State \(\bm{x} \gets \text{Decoder}(\bm{z}_0)\)
\State \textbf{return} \(\bm{x}\)
\end{algorithmic}
\end{algorithm}

\section{Experiments}
\label{exp}
\subsection{Implementation Details}
\label{implementation}

Both PixelINN and LatentINN ($\approx0.73/0.74 \mathrm{M}$ parameters) adopt a two-level design (i.e., two InvBlocks). Within each block, split/merge operations are performed via the wavelet transform: PixelINN uses a non-redundant Haar wavelet transform, whereas LatentINN employs a redundant undecimated Haar transform.
Moreover, each InvBlock contains two pairs of PM–UM modules, where each PM or UM module consists of two MRSTBs, each with two STLs.
For degradation estimation, we directly adopt the pre-trained implicit degradation estimator $\mathcal{F}(\cdot)$ in \cite{kdsr} and generate $\bm{\gamma}_{\mathrm{deg}} = \mathcal{F}(\bm{y})$. We train our INNs with the Adam optimizer ($\beta_1=0.9$, $\beta_2=0.999$), batch size $8$, and learning rate $5\times 10^{-5}$.

For blind face restoration, we train our INNs on the FFHQ \cite{ffhq} dataset and evaluate these models on a synthetic dataset, CelebA-Test \cite{celeba}, as well as two real-world datasets: WebPhoto-Test \cite{GFPGAN}, and CelebChild \cite{GFPGAN}. Specifically, WebPhoto-Test \cite{GFPGAN} consists of 407 face images from low-quality photos
in real life from the Internet, and CelebChild contains 180 child celebrity faces collected from the Internet \cite{GFPGAN}.
{{To further explore the generalization capability of our proposed method, we also perform experiments on natural image restoration tasks on DIV2K \cite{div2k}, RealSR \cite{realsr}, and DRealSR \cite{drealsr} datasets.}}
The reconstruction results are evaluated with PSNR, LPIPS \cite{lpips}, DISTS \cite{dists}, FID \cite{fid}, PI \cite{blau2018perception}, NRQM \cite{Ma2017NRQM}, DBCNN \cite{Zhang2020DBCNN}, and CNNIQA \cite{Kang2014CNNIQA}.

\vspace{-0.2cm}
\subsection{Results on Blind Image Restoration}
In this section, we perform quantitative and qualitative assessments of the results produced by the proposed LatentINDIGO and other state-of-the-art methods and then discuss the improvements brought by our approaches.

\subsubsection{Results on Blind Face Restoration}

\begin{table}[]\footnotesize
\centering
\caption{Quantitative comparison on \textit{CelebA-Test}. The best and second best results are highlighted in \textcolor{red}{\textbf{red}} and \textcolor{blue}{\underline{blue}}, respectively.}
\resizebox{0.48\textwidth}{!}{
\begin{tabular}{@{}c|c|ccccc@{}}
\hline
Degradation &{Method} & PSNR $\uparrow$ & LPIPS $\downarrow$ & DISTS $\downarrow$ & FID $\downarrow$ \\ \hline
\multirow{8}{*}{\begin{tabular}[c]{@{}c@{}}\textit{Mild}\end{tabular}} &PGDiff & 23.17 & 0.2691 & 0.1595 &25.95 \\
 &Difface &24.78  & 0.2578 & 0.1584 & 22.30 \\
 &DR2       &{25.72}  & 0.2745 & 0.1814 & 28.70  \\
 &SeeSR   & 25.04  & 0.2447 & 0.1590 & 28.23 \\ 
 &StableSR   & 24.92  & 0.2261 & {0.1465} & \second{20.08} \\
 &DiffBIR   & 25.46  & 0.2277 & 0.1575& 27.21 \\

&\mygraycell\textbf{DiffBIR-LatentINN}                            & \mygraycell \second{26.01}  &\mygraycell \best{0.2180} & \mygraycell\best{0.1434} &\mygraycell 20.19  \\  
  & \mygraycell\textbf{DiffBIR-PixelINN}   & \mygraycell\best{26.68}  & \mygraycell\second{0.2192} & \mygraycell0.1585 &\mygraycell25.72 \\
 & \mygraycell\textbf{StableSR-PixelINN}   & \mygraycell25.20 & \mygraycell {0.2222} & \mygraycell\second{0.1456} & \mygraycell\best{19.83} \\\hline
\multirow{10}{*}{\begin{tabular}[c]{@{}c@{}}\textit{Medium}\end{tabular}}   &PGDiff & 22.42  & 0.2894 & 0.1665 & 29.68 \\
 &Difface   & 24.30  & 0.2734 & 0.1678 & 23.63 \\
  & DR2        & 24.09  & 0.2924 & 0.1966 & 33.87 \\
 &SeeSR         & 23.60 & 0.2834 & 0.1708 & 31.39 \\
 &StableSR       & 23.69  & 0.2566 & \second{0.1564} & \second{22.49} \\
 &DiffBIR                            & {24.41}  & 0.2602 & 0.1689 & 29.84 \\ 
     &\mygraycell \textbf{DiffBIR-LatentINN}       & \mygraycell\second{25.07} & \mygraycell{0.2490} & \mygraycell0.1570 & \mygraycell22.89 \\
    &\mygraycell \textbf{DiffBIR-PixelINN}       & \mygraycell\best{25.18} & \mygraycell\best{0.2456} & \mygraycell0.1639 & \mygraycell25.12 \\
   &\mygraycell \textbf{StableSR-PixelINN}       &\mygraycell 24.15  & \mygraycell\second{0.2480} & \mygraycell\best{0.1552} &\mygraycell \best{21.40} \\\hline
 \multirow{9}{*}{\begin{tabular}[c]{@{}c@{}}\textit{Severe}\end{tabular}}&PGDiff  & 21.82  & 0.3085 & 0.1727 &33.18 \\
 &DifFace          &23.52 & 0.3043 & 0.1853 & 28.31 \\
 &DR2         & 23.22 & 0.3027 & 0.1942 &32.77 \\
 &SeeSR                            & 22.73 & 0.3087 & 0.1790 & 33.29 \\ 
&StableSR      & 22.80& \second{0.2791} & \second{0.1630} & \second{23.46}\\
 &DiffBIR      & {23.74}  &0.3054 & 0.1866 & 32.59\\
&\mygraycell \textbf{DiffBIR-LatentINN}       & \mygraycell\second{24.16} & \mygraycell{0.2951} & \mygraycell0.1795 & \mygraycell29.76 \\
&\mygraycell\textbf{DiffBIR-PixelINN}                            & \mygraycell \best{24.27}  &\mygraycell 0.2903 & \mygraycell0.1822 &\mygraycell 28.32  \\  
&\mygraycell\textbf{StableSR-PixelINN}                            & \mygraycell23.23& \mygraycell\best{0.2739} & \mygraycell \best{0.1619} & \mygraycell\best{23.10} \\\hline 
 
\end{tabular}
}
\label{tab:celeba}
\end{table}

\begin{table}[]\footnotesize
\centering
\caption{Quantitative comparison on real-world datasets. The best and second best results are highlighted in \textcolor{red}{\textbf{red}} and \textcolor{blue}{\underline{blue}}, respectively.}
\resizebox{0.48\textwidth}{!}{
\begin{tabular}{@{}c|c|ccccc@{}}
\hline
Dataset &{Method} &{         PI $\downarrow$ }& NRQM $\uparrow$ & DBCNN $\uparrow$ &CNNIQA $\uparrow$ \\ \hline

\multirow{8}{*}{\begin{tabular}[c]{@{}c@{}}\textit{WebPhoto}\end{tabular}}   &PGDiff & \second{4.0209}	&6.6095	&0.5570&	0.5395\\
 &Difface   & 4.8747&	5.4389&	0.5333&	0.5056 \\
  & DR2        & 6.2217	&4.0648	&0.4970	&0.4822 \\
 &SeeSR         & 4.8803&	6.2008&	0.5925	&0.5797\\
 &StableSR       & 4.2030&	\second{6.8831}	&0.5952&	0.5646\\
 &DiffBIR                            & 4.8814	&6.8515&	0.6160&0.5911\\ 
   &\mygraycell \textbf{DiffBIR-LatentINN}       & \mygraycell 4.8700 & \mygraycell6.8680 & \mygraycell \second{0.6171} & \mygraycell\second{0.5924}\\
  &\mygraycell \textbf{DiffBIR-PixelINN}       & \mygraycell\best{3.8335} & \mygraycell\best{7.8747} & \mygraycell\best{0.6526} & \mygraycell\best{0.6386}\\
\hline

 \multirow{8}{*}{\begin{tabular}[c]{@{}c@{}}\textit{Child}\end{tabular}}&PGDiff  & 3.4668	& 7.4635 &	0.6076&	0.5966\\
 &DifFace          &4.2303&	6.3257	&0.5443	&0.5185 \\
 &SeeSR                            & 4.0003	&7.3328	&0.6348	&\second{0.6325} \\ 
&StableSR      & \second{3.4580}	&\second{7.7222}&	0.5854	&0.5714\\
 &DiffBIR      & 4.0712&	7.5967&	{0.6357}	&{0.6205}\\
 &\mygraycell\textbf{DiffBIR-LatentINN}                            & \mygraycell {4.0507}  &\mygraycell{7.6164} & \mygraycell\second{0.6407} &\mygraycell {0.6247} \\  
&\mygraycell\textbf{DiffBIR-PixelINN}                            & \mygraycell \best{3.3063}  &\mygraycell\best{8.0433} & \mygraycell\best{0.6694} &\mygraycell\best{0.6572} \\  
\hline

\end{tabular}
}
\label{tab:real_v2}
\end{table}

\begin{figure*}[!tp]\footnotesize 
	\centering
\hspace{-0.2cm}
\begin{tabular}{c@{\extracolsep{0.1em}}c@{\extracolsep{0.1em}}c@{\extracolsep{0.1em}}c@{\extracolsep{0.1em}}c@{\extracolsep{0.1em}}c@{\extracolsep{0.1em}}c@{\extracolsep{0.1em}}c@{\extracolsep{0.1em}}c@{\extracolsep{0.1em}}c@{\extracolsep{0.1em}}c}
                &\includegraphics[width=0.1\textwidth]{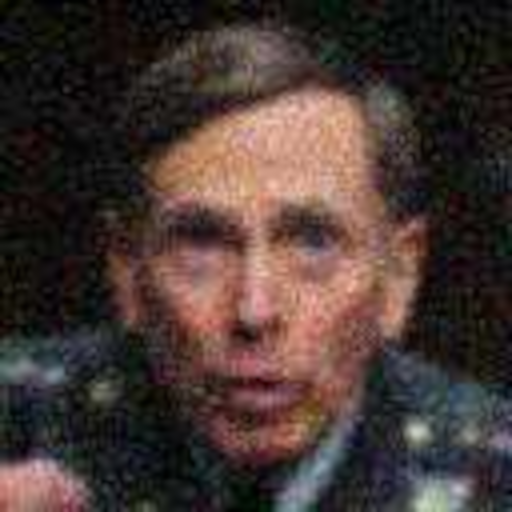}~
		&\includegraphics[width=0.1\textwidth]{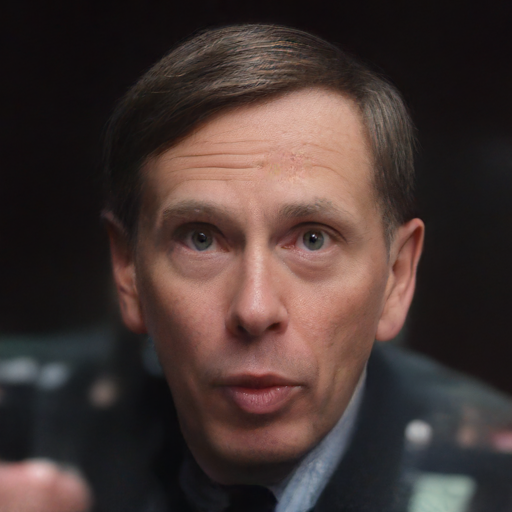}~
    &\includegraphics[width=0.1\textwidth]{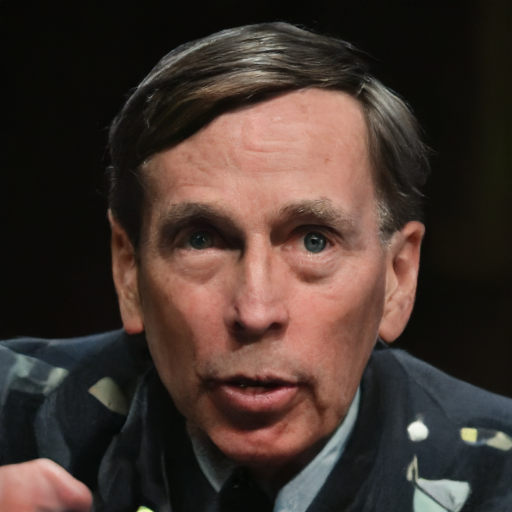}~
    &\includegraphics[width=0.1\textwidth]{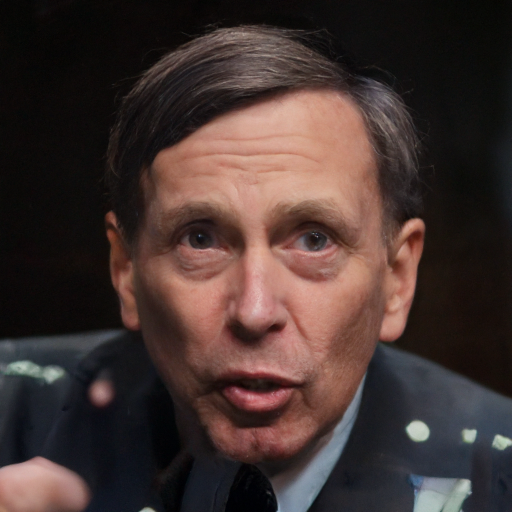}~
    &\includegraphics[width=0.1\textwidth]{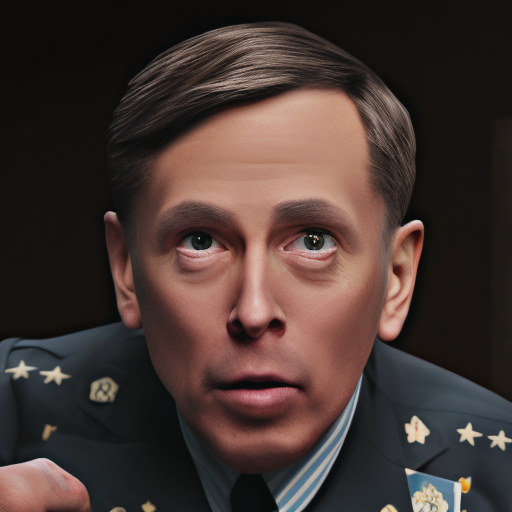}~
        &\includegraphics[width=0.1\textwidth]{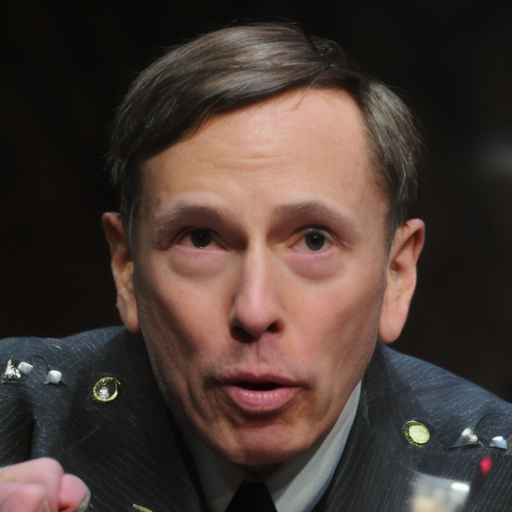}~
        &\includegraphics[width=0.1\textwidth]{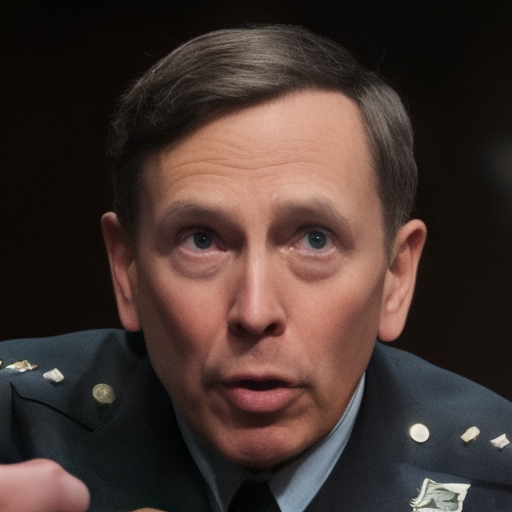}~
		&\includegraphics[width=0.1\textwidth]{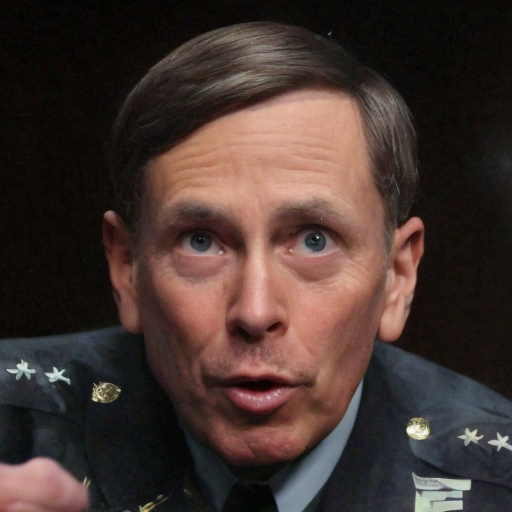}~
		&\includegraphics[width=0.1\textwidth]{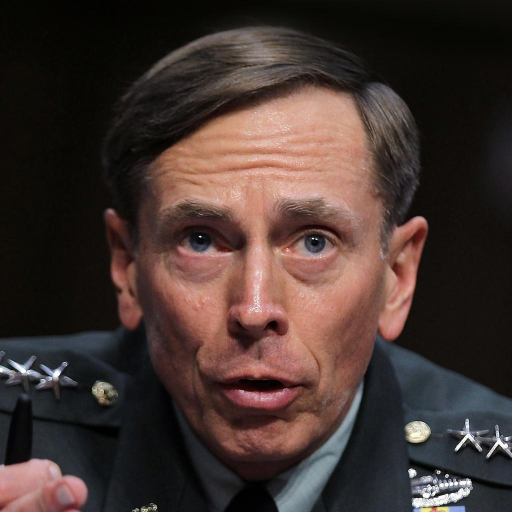}\\
            &\includegraphics[width=0.1\textwidth]{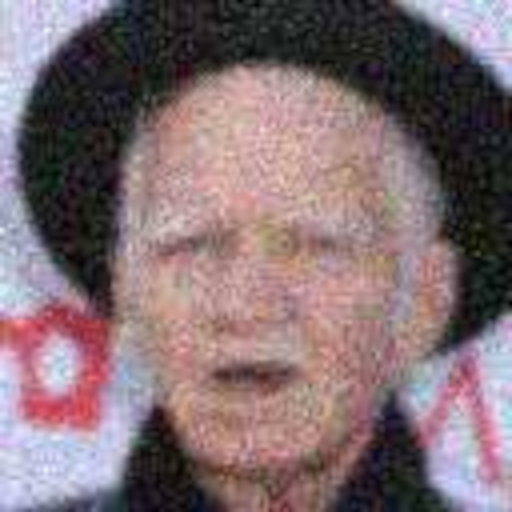}~
		&\includegraphics[width=0.1\textwidth]{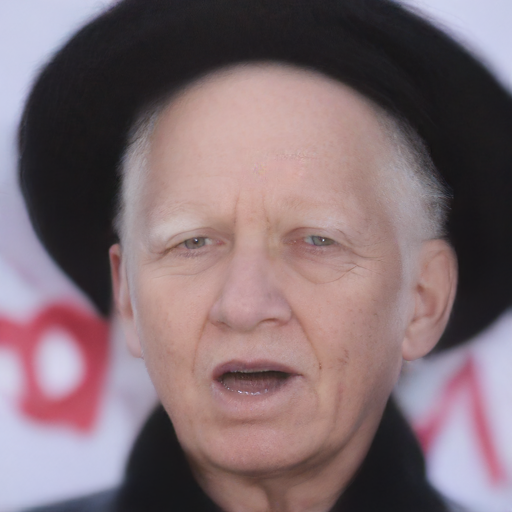}~
    &\includegraphics[width=0.1\textwidth]{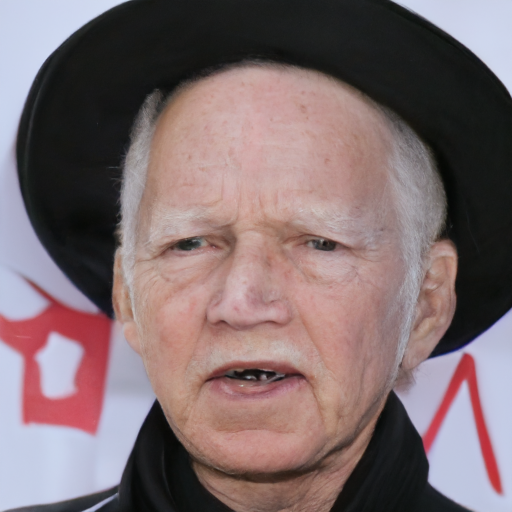}~
    &\includegraphics[width=0.1\textwidth]{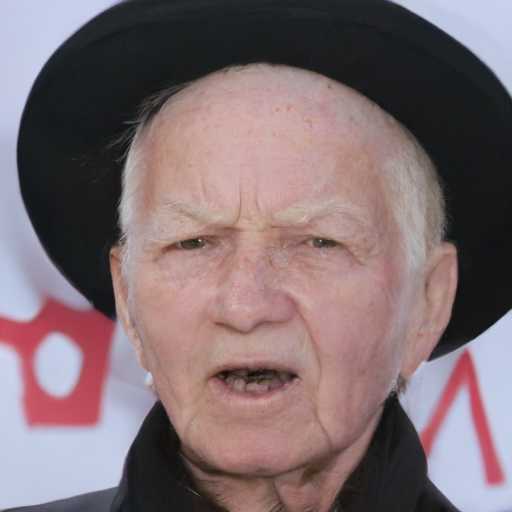}~
    &\includegraphics[width=0.1\textwidth]{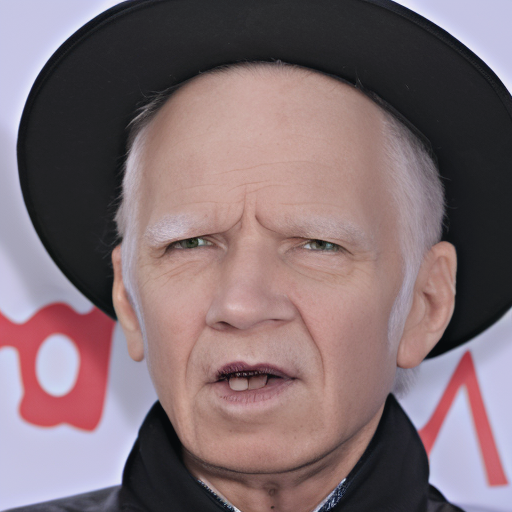}~
        &\includegraphics[width=0.1\textwidth]{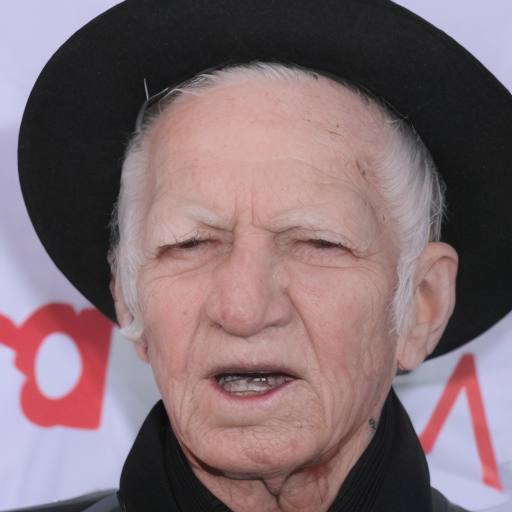}~
        &\includegraphics[width=0.1\textwidth]{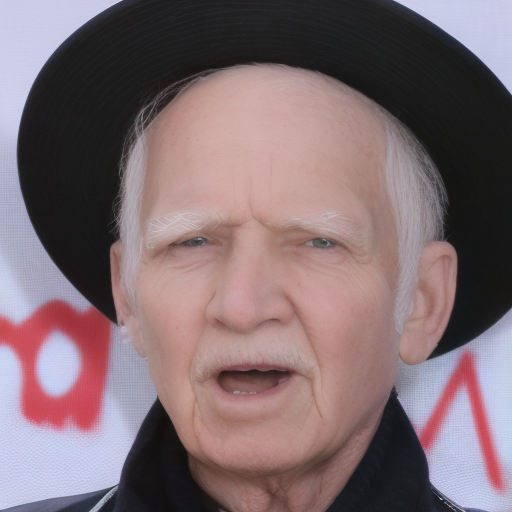}~
		&\includegraphics[width=0.1\textwidth]{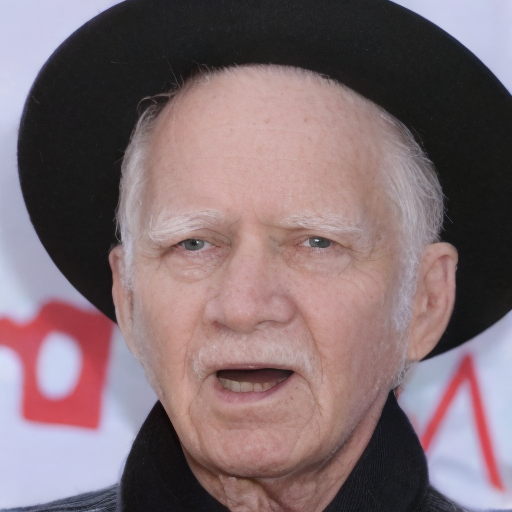}~
		&\includegraphics[width=0.1\textwidth]{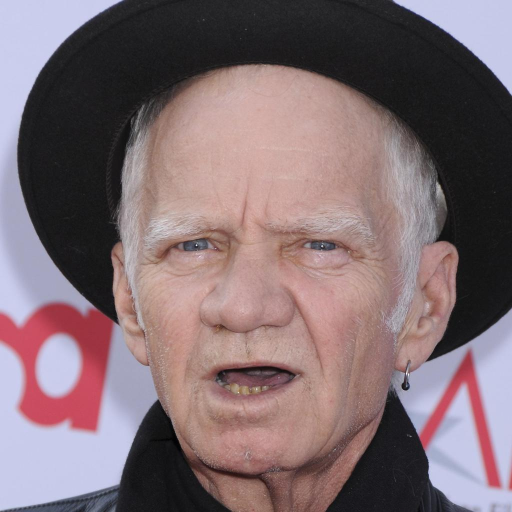}\\
&Input&DR2 &PGDiff &DifFace &SeeSR &StableSR &DiffBIR &\makecell[c]{\textbf{DiffBIR}\\\textbf{-LatentINN}}&Ground-Truth \\
    
	\end{tabular}
    \vspace{-0.3cm}
	\caption{{Comparisons on 4x {blind} SR with medium degradation on CelebA-HQ.}} 
	\label{fig:medium}
\end{figure*}

\begin{figure*}[!tp]\footnotesize 
	\centering
\hspace{-0.2cm}
\begin{tabular}{c@{\extracolsep{0.1em}}c@{\extracolsep{0.1em}}c@{\extracolsep{0.1em}}c@{\extracolsep{0.1em}}c@{\extracolsep{0.1em}}c@{\extracolsep{0.1em}}c@{\extracolsep{0.1em}}c@{\extracolsep{0.1em}}c}
    &\includegraphics[width=0.115\textwidth]{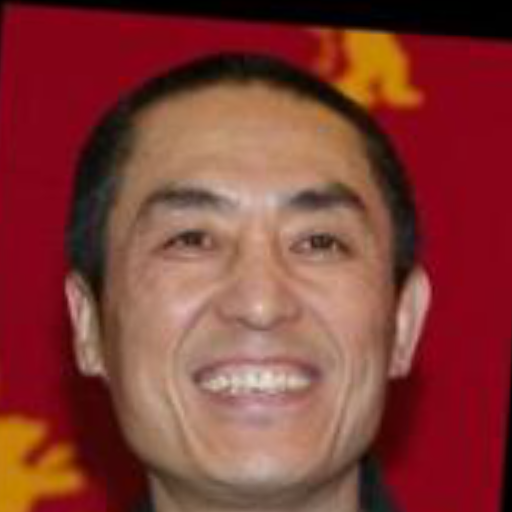}~
		&\includegraphics[width=0.115\textwidth]{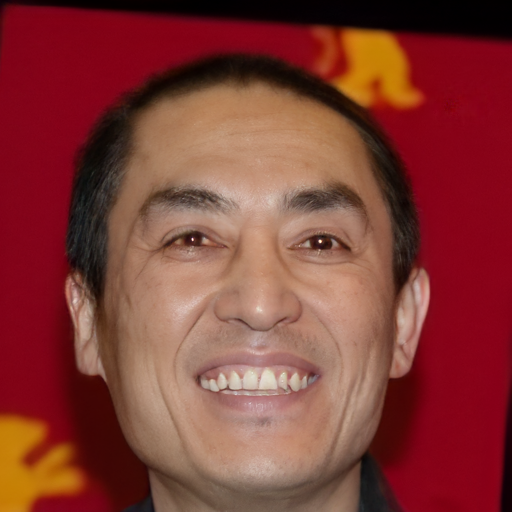}~

    &\includegraphics[width=0.115\textwidth]{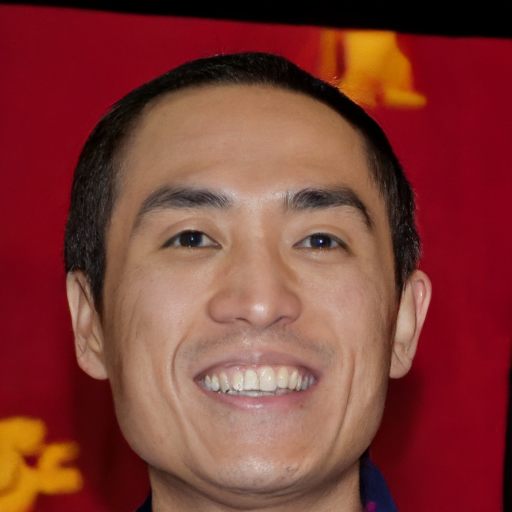}
    ~&\includegraphics[width=0.115\textwidth]{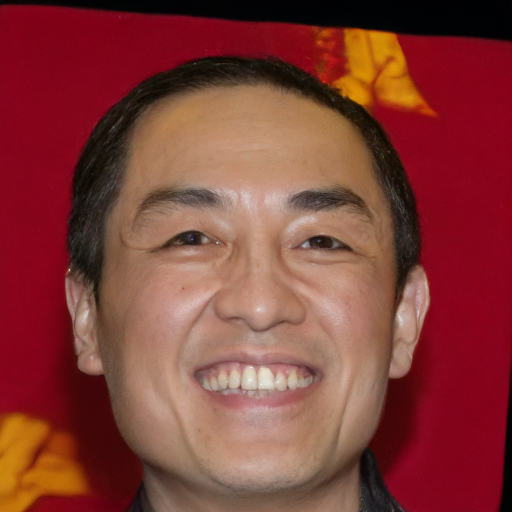}~

    &\includegraphics[width=0.115\textwidth]{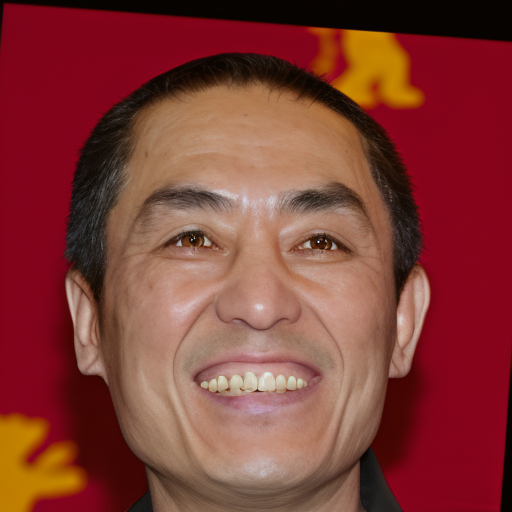}~
        &\includegraphics[width=0.115\textwidth]{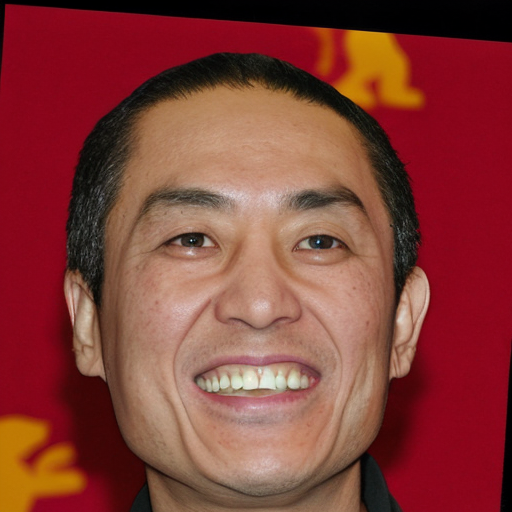}~
        &\includegraphics[width=0.115\textwidth]{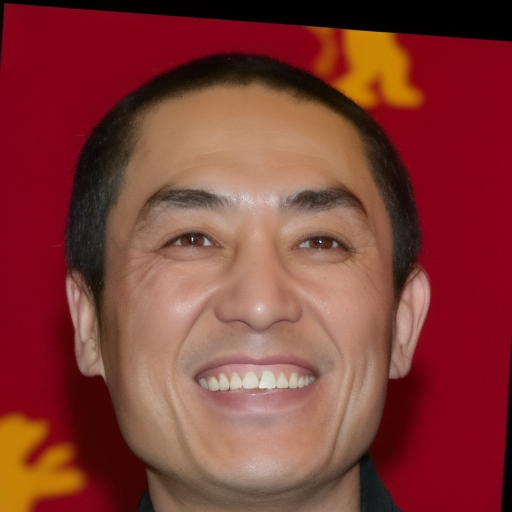}~
		&\includegraphics[width=0.115\textwidth]{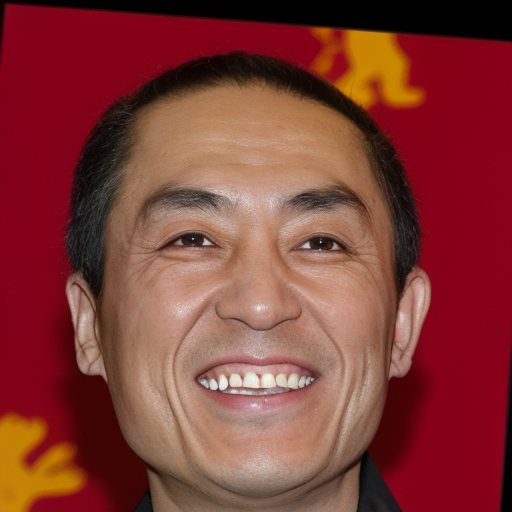}\\
            &\includegraphics[width=0.115\textwidth]{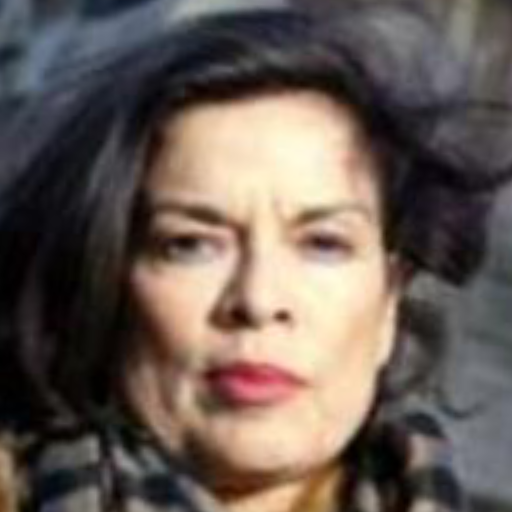}~
		&\includegraphics[width=0.115\textwidth]{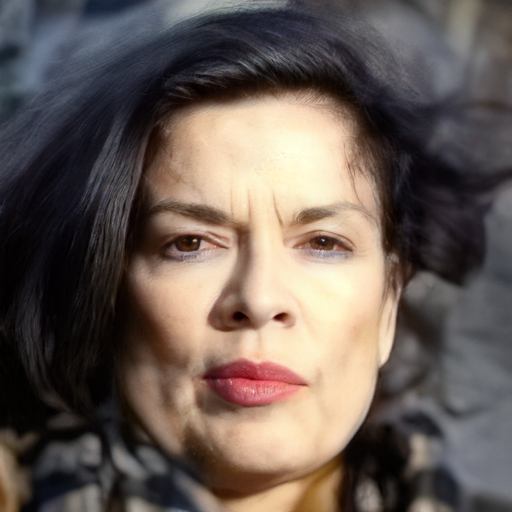}~

    &\includegraphics[width=0.115\textwidth]{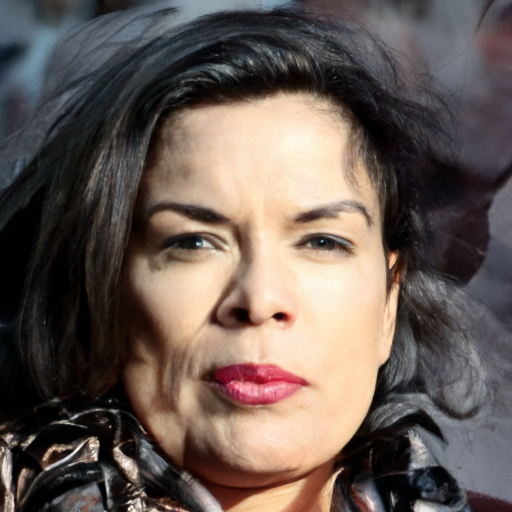}
    ~&\includegraphics[width=0.115\textwidth]{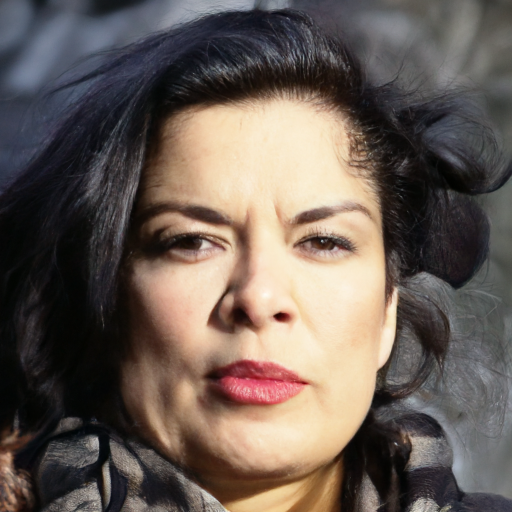}~

    &\includegraphics[width=0.115\textwidth]{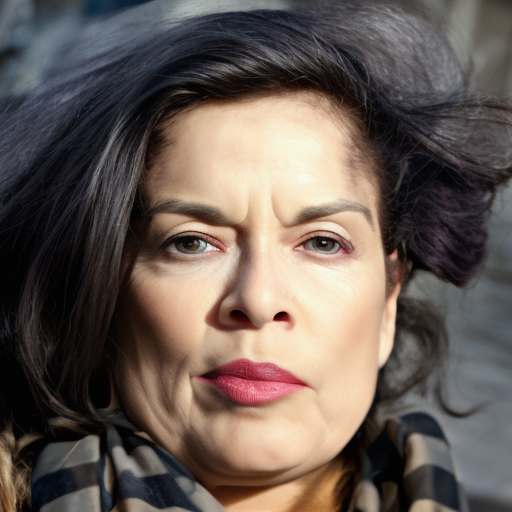}~
        &\includegraphics[width=0.115\textwidth]{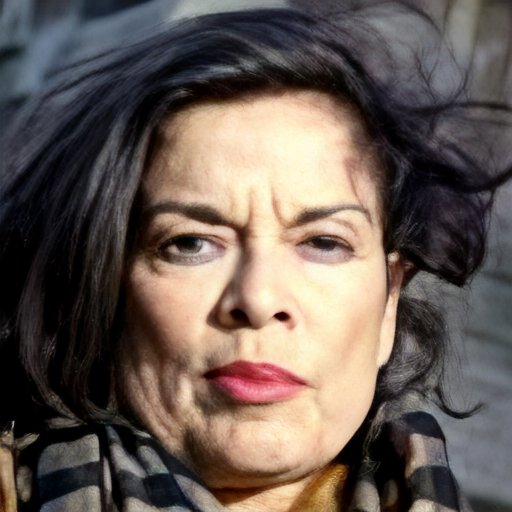}~
        &\includegraphics[width=0.115\textwidth]{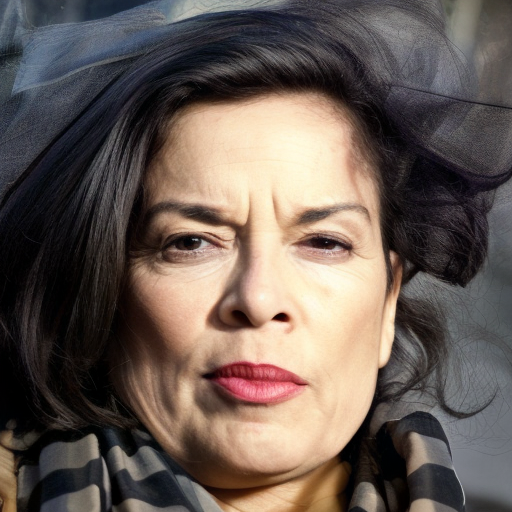}~
		&\includegraphics[width=0.115\textwidth]{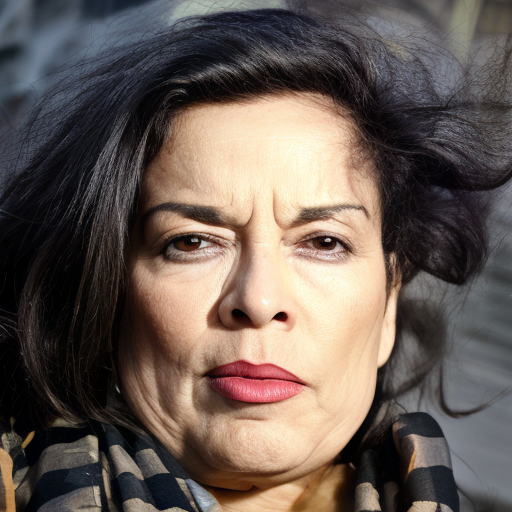}\\
&Input&DR2&PGDiff&DifFace&SeeSR&StableSR&DiffBIR&\makecell[c]{\textbf{DiffBIR}\\\textbf{-LatentINN}}\\
    
	\end{tabular}
    \vspace{-0.2cm}
	\caption{{Comparisons on reconstruction results on real-world datasets.}} 
	\label{fig:lfw}
\end{figure*}

\begin{figure*}[!tp]\footnotesize 
	\centering
\hspace{-0.2cm}
\begin{tabular}{c@{\extracolsep{0.12em}}c@{\extracolsep{0.12em}}c@{\extracolsep{0.12em}}c@{\extracolsep{0.12em}}c@{\extracolsep{0.12em}}c}
                        &\includegraphics[width=0.12\textwidth]{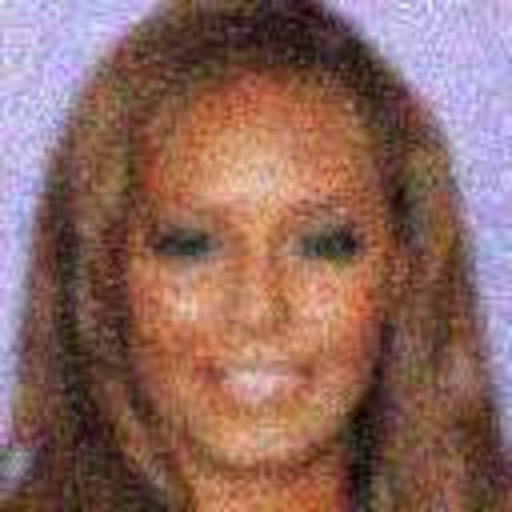}~
                &\includegraphics[width=0.12\textwidth]{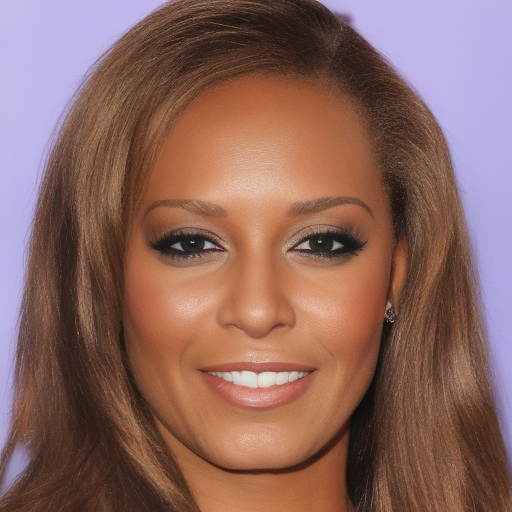}~
    &\includegraphics[width=0.12\textwidth]{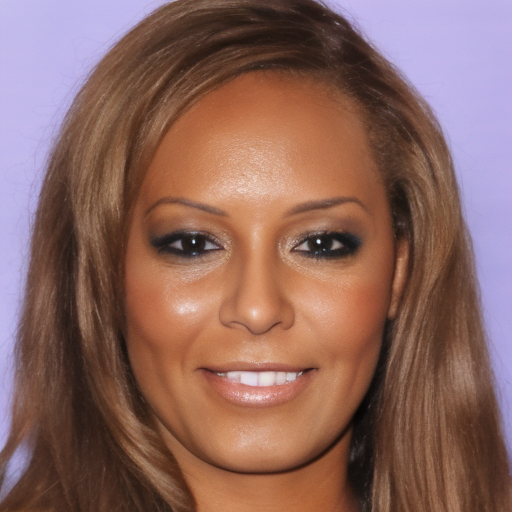}~
    &\includegraphics[width=0.12\textwidth]{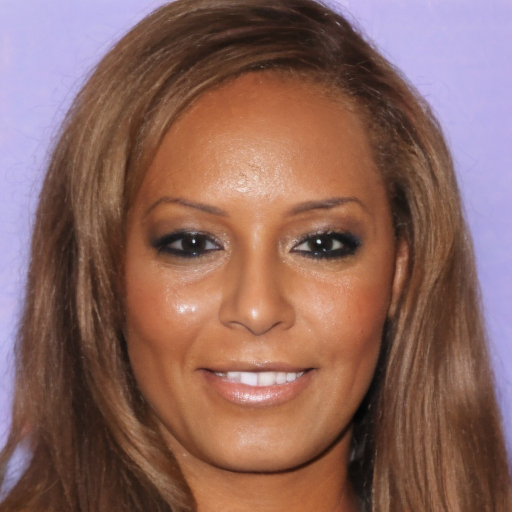}~
        &\includegraphics[width=0.12\textwidth]{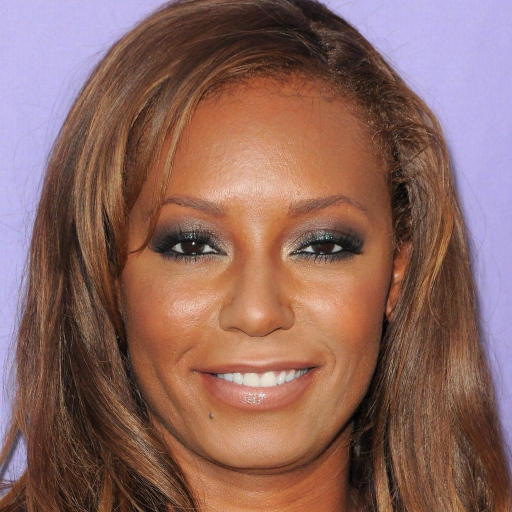}\\
        &\includegraphics[width=0.12\textwidth]{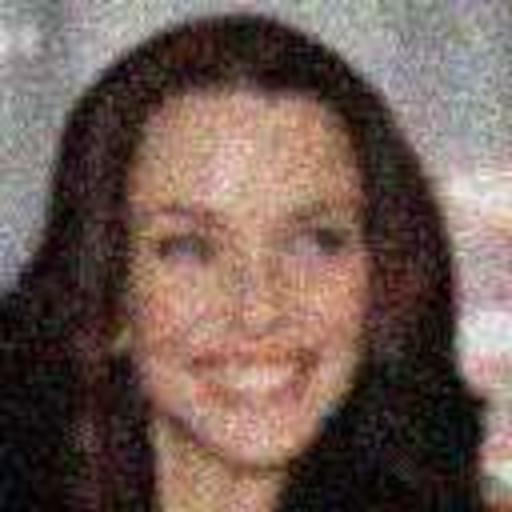}~
    &\includegraphics[width=0.12\textwidth]{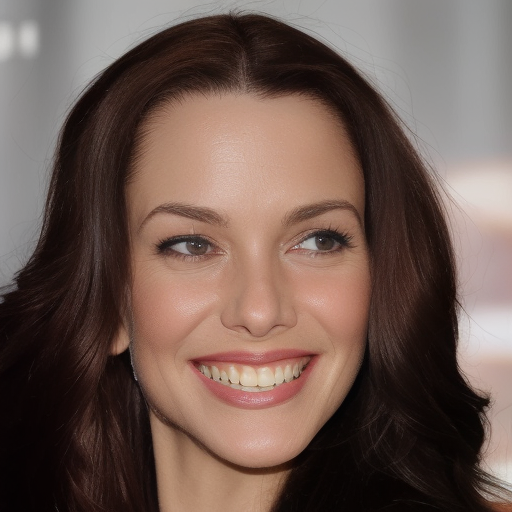}~
    &\includegraphics[width=0.12\textwidth]{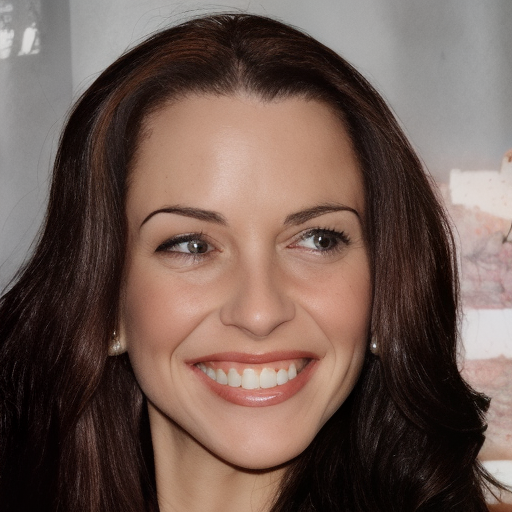}~
    &\includegraphics[width=0.12\textwidth]{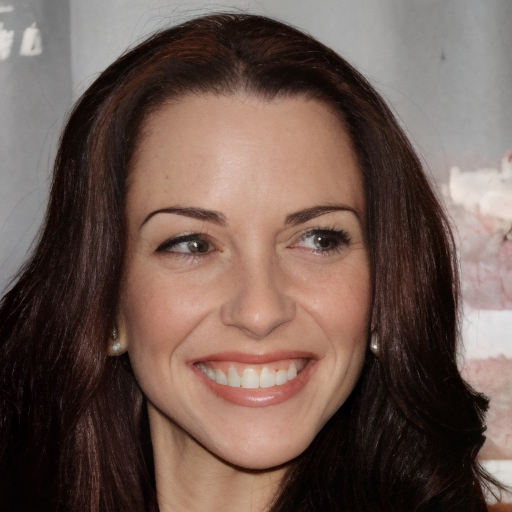}~
        &\includegraphics[width=0.12\textwidth]{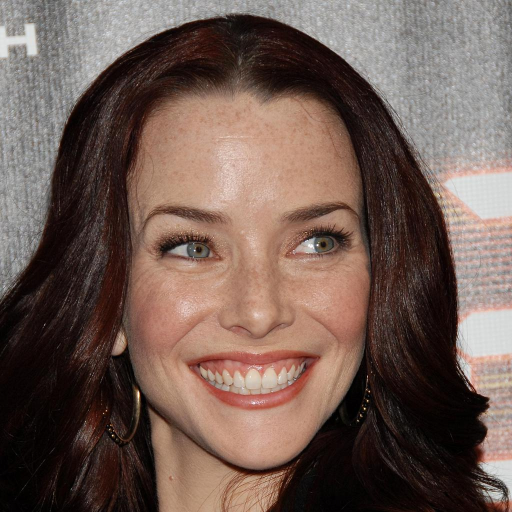}\\

&Input&DiffBIR&\textbf{DiffBIR-PixelINN}&\textbf{DiffBIR-LatentINN}&Ground-Truth\\
    
	\end{tabular}
    \vspace{-0.1cm}
	\caption{{Visual comparison on \(4\times\) blind super-resolution among baseline DiffBIR, and our proposed LatentINDIGO-PixelINN and LatentINDIGO-LatentINN.  }} 
 \vspace{-0.3cm}
	\label{fig:pinn_vs_linn}
\end{figure*}

We compare our approach with several state-of-the-art blind
face restoration methods: PGDiff \cite{yang2023pgdiff}, DifFace \cite{difface}, DR2 \cite{wang2023dr2}, SeeSR \cite{wu2024seesr}, StableSR \cite{wang2024exploiting} and DiffBIR \cite{lin2023diffbir}. Specifically, we adopt StableSR and DiffBIR as baseline methods, reusing their pre-trained LDM denoiser \(\epsilon_{\theta_{IR}}(\bm{z}_t, t, \bm{y})\) in line 3 of Algorithm~\ref{alg:indigo_latent} (with PixelINN) and Algorithm~\ref{alg:latent} (with LatentINN). (More results of our approach with the unconditional pre-trained denoiser \(\epsilon_{\theta}(\bm{z}_t, t)\) of LDM \cite{rombach2022high} can be found in supplementary material.) In our implementation for this case, we train both INNs on data degraded by: $\bm{y} = \left[(\bm{x}
 \circledast
 \bm{k}_{\sigma})_{\downarrow_{r}} + \bm{n}_{\delta}\right]_{\mathtt{JPEG}_{q}}$, where \( \bm{x} \) is the high-quality image, \( \circledast \) denotes convolution, \( \bm{k}_{\sigma} \) represents the Gaussian blurring kernel, \( \downarrow_{r} \) indicates downsampling with scale factor \( r \), \( \bm{n}_{\delta} \) is the additive white Gaussian noise, and \( \mathtt{JPEG}_{q} \) denotes JPEG compression with quality factor \( q \), where we set $r = 4$ and randomly sample $\sigma, \delta,$ and $q$ from the intervals $[3,9]$, $[5,50]$, and $[30,80]$, respectively. To evaluate the effectiveness of our framework, we implement three variants, StableSR-PixelINN, DiffBIR-PixelINN, and DiffBIR-LatentINN, named to reflect both the baseline LDM denoiser (StableSR or DiffBIR) and the type of our approach (PixelINN or LatentINN).

\textbf{Synthetic CelebA-Test.} We evaluate our approach using the CelebA HQ 512×512 1k validation dataset \cite{celeba} on synthetic degradation. To evaluate these methods on different levels of degradation, we test them on mild ($\sigma$=4, $\delta$=15, $q$=70), medium ($\sigma$=6, $\delta$=25, $q$=50), and severe ($\sigma$=8, $\delta$=35, $q$=30) degradations, respectively. We provide the quantitative comparison on different levels of degradations in Table \ref{tab:celeba}. When comparing our approaches with the baseline models, we find that our LatentINDIGO achieves consistent improvements on all evaluation metrics, without requiring any retraining or finetuning of the LDMs. In particular, DiffBIR-PixelINN achieves improvements of up to \(1.22\) dB in PSNR and reduces LPIPS by up to \(0.0151\), while DiffBIR-LatentINN attains up to \(0.66\) dB improvement in PSNR and lowers LPIPS by \(0.0112\). Furthermore, in comparison to other state-of-the-art methods, our approaches achieve the best performance under all three degradation settings. Qualitative results in Fig.~\ref{fig:medium} also demonstrate the superiority of LatentINDIGO, revealing that its reconstructions align more closely with the ground truth and present more realistic texture details than those of competing methods. (More results are provided in the supplementary material.)
Fig.~\ref{fig:pinn_vs_linn} compares the performance of DiffBIR-PixelINN and DiffBIR-LatentINN. Both methods deliver high-quality reconstructions but exhibit distinct strengths: PixelINN often produces globally consistent appearances, whereas LatentINN captures finer details, such as wrinkles and facial shine.

\textbf{Real-World WebPhoto-Test and CelebChild.} To test the generalization ability, we evaluate our framework on two real-world datasets: WebPhoto-Test \cite{GFPGAN}, and CelebChild \cite{GFPGAN}. As shown in Table \ref{tab:real_v2}, we report PI \cite{blau2018perception}, NRQM \cite{Ma2017NRQM}, DBCNN \cite{Zhang2020DBCNN}, and CNNIQA \cite{Kang2014CNNIQA} scores across these datasets and present comprehensive quantitative results. By comparing our approaches with the baseline, DiffBIR, we observe substantial improvements on both our LatentINDIGO-LatentINN and LatentINDIGO-PixelINN, establishing state-of-the-art performance on all tested datasets. Furthermore, a qualitative comparison illustrated in Fig.~\ref{fig:lfw}, further demonstrates the superior restoration capability of our proposed method.

\subsubsection{Results on BIR on Natural Images}
\label{natural images}

\begin{figure}[!tp]\footnotesize 
	\centering
\hspace{-0.2cm}
\begin{tabular}{c@{\extracolsep{0.1em}}c@{\extracolsep{0.1em}}c@{\extracolsep{0.1em}}c@{\extracolsep{0.1em}}c}

    &\includegraphics[width=0.11\textwidth]{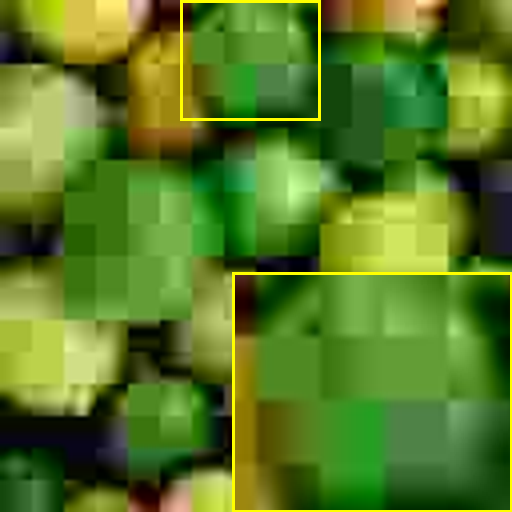}~
    &\includegraphics[width=0.11\textwidth]{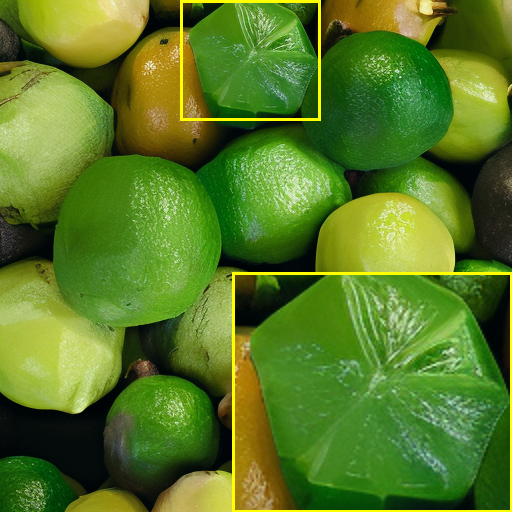}~
    &\includegraphics[width=0.11\textwidth]{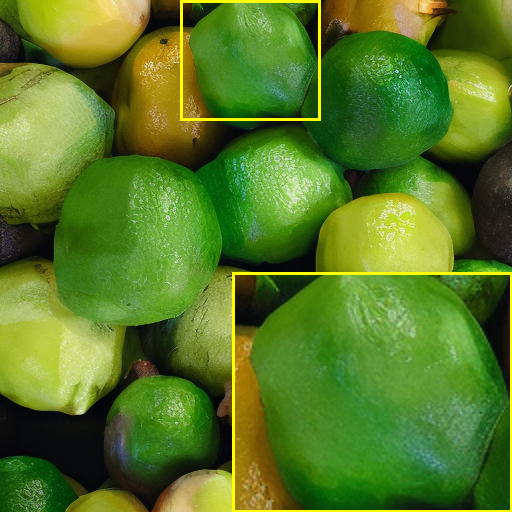}
    ~&\includegraphics[width=0.11\textwidth]{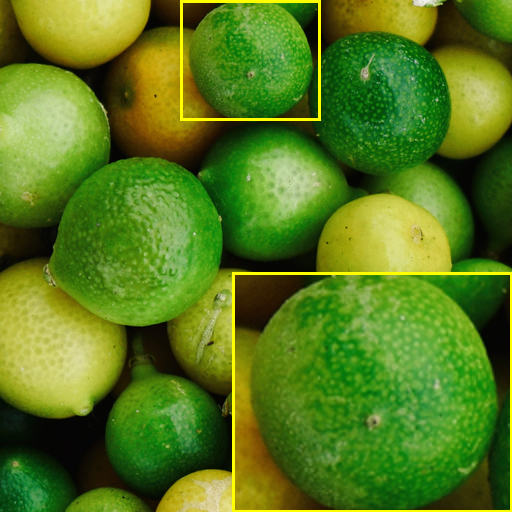}\\
&Input&SeeSR&\textbf{SeeSR-PixelINN}&Ground-Truth\\
    
	\end{tabular}
	\caption{{Comparisons on JPEG compression (q=5) on DIV2K dataset \cite{div2k}.}} 
	\label{fig:jpeg}
\end{figure}

\begin{figure}[t]
  \centering
    \captionsetup[sub]{font=tiny}
  \subfloat[Input image from DRealSR Dataset \cite{drealsr}. Input size: 64$\times$256]{
    \includegraphics[width=0.86\linewidth]{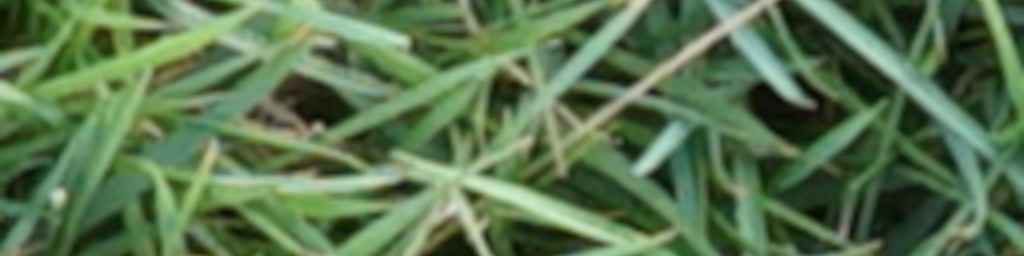}

  }
  \vspace{0.1cm}
  \subfloat[Our result. Output size: 256$\times$1024.]{
    \includegraphics[width=0.86\linewidth]{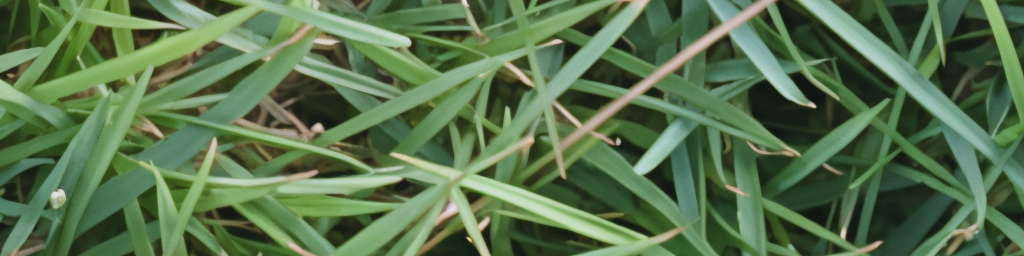}
  }
  \caption{Our results on arbitrary-size BIR. Subfigures show the input and output sizes.}
  \label{fig:arbi_size}
  \vspace{-0.5cm}
\end{figure}

To implement our LatentINDIGO on natural images, we train a PixelINN on the DIV2K~\cite{div2k} training set with synthetic degradation. We adopt SeeSR~\cite{wu2024seesr} as our baseline, using its pre-trained \(\epsilon_{\theta_{IR}}(\bm{z}_t, t, \bm{y}, \bm{\gamma}_{test})\) in line 3 of Algorithm~\ref{alg:latent}.
To evaluate our approach on a degradation not consistent with the training pipeline of our INNs, we test our approach on an unseen JPEG compression (q=5) with only our refinement mechanism (rather than retraining an INN from scratch). As shown in Fig. \ref{fig:jpeg}, the proposed LatentINDIGO surpasses the baseline SeeSR and further demonstrates superior flexibility.
Finally, Fig. \ref{fig:arbi_size} illustrates the capability of our method to support arbitrary-size reconstruction, highlighting its flexibility and robustness in handling diverse input resolutions. Further comparisons and discussions can be found in supplementary material.

\vspace{-0.3cm}
\subsection{Analysis}
\label{ana}
\subsubsection{Comparison of Guidance Strategies for IR with LDM}
\label{compare_guidance}

In this section, we investigate the impact of different guidance strategies on a shared pretrained latent diffusion baseline and compare them with our proposed approach. We begin by examining the LDPS family~\cite{psld,p2l,mpgd,resample,treg,ldir}, which enforces data-consistency during sampling through the objective
\(
\|\bm{y} - \mathcal{H}\mathcal{D}(\bm{z}_{0,t})\|_2^2,
\)
where \(\mathcal{H}\) is a known degradation operator, and \(\mathcal{D}\) maps the latent variable \(\bm{z}_{0,t}\) to the pixel space.
To align LDPS with our \textit{blind} framework, we approximate the degradation operator with the forward transform of our PixelINN, and replace our PixelINN-guidance update with a gradient step that enforces LDPS-style data consistency:
\begin{equation}
      \vspace{-0.19cm}
\tilde{\bm{z}}_{0,t} \;=\; \bm{z}_{0,t} \;-\; \zeta \,\nabla_{\bm{z}_{0,t}}
\|\bm{x}_{c,t} - \bm{y}\|_2^2,
\label{eq:latent_dps_modification}
\end{equation}
where \(\bm{x}_{c,t}\) is generated from
$
[\bm{x}_{c,t};\,\bm{x}_{d,t}]
= g_{\mathbf{\Theta}_{pix}}\bigl(\mathcal{D}(\bm{z}_{0,t}),\bm{\gamma}_{deg}\bigr),
$
as specified in our algorithm. In this manner, \(\mathcal{H}\) is replaced by the learned operator \(g_{\mathbf{\Theta}_{pix}}\), thereby enabling LDPS to operate under unknown forward processes, for further comparison.

The second approach, PGDiff~\cite{yang2023pgdiff}, originally developed for pixel-domain diffusion models, employs the restored output of a standard MSE-based inverse restoration (IR) network, \({f}_{\mathbf{\Theta}_\text{IR}}(\bm{y})\), to guide the sampling process, thereby mitigating hallucination and improving fidelity. This strategy has also been adopted by recent LDM-based methods~\cite{wang2024exploiting,lin2023diffbir}, wherein minimizing
\(\|{f}_{\mathbf{\Theta}_\text{IR}}(\bm{y}) - \mathcal{D}(\bm{z}_{0,t})\|_2^2\)
during sampling to balance fidelity and perceptual quality.
To incorporate PGDiff into our framework, we adopt the widely used SwinIR \cite{liang2021swinir} as \(f_{\mathbf{\Theta}_\text{IR}}(\cdot)\) and implement the following gradient update:
\begin{equation}
\tilde{\bm{z}}_{0,t} \;=\; \bm{z}_{0,t} \;-\; \zeta\,\nabla_{\bm{z}_{0,t}}
\bigl\|\mathcal{D}(\bm{z}_{0,t}) - {f}_{\mathbf{\Theta}_\text{IR}}(\bm{y})\bigr\|_2^2.
\label{eq:latent_pgdiff_modification}
\end{equation}

The third approach, {High-Frequency Guidance Sampling (HGS)}, as introduced by PromptFix~\cite{promptfix}, aims to preserve high-frequency details by employing high-pass operators \(\Phi_\text{HP}\) (e.g., the Sobel operator~\cite{pratt2007digit}). This is accomplished by minimizing the discrepancy between the high-frequency components of the observed data \(\bm{y}\) and those of the reconstructed output \(\mathcal{D}(\bm{z}_{0,t})\), expressed as
\(
\|\Phi_\text{HP}(\bm{y}) - \Phi_\text{HP}(\mathcal{D}(\bm{z}_{0,t}))\|_2^2.
\)
Because our proposed method and HGS rely on different baseline LDM
models, we incorporate HGS into our framework by replacing our PixelINN guidance step in Algorithm~\ref{alg:indigo_latent} with their high-frequency guidance update:
\begin{equation}
\tilde{\bm{z}}_{0,t}
= \bm{z}_{0,t}
- \zeta \,\nabla_{\bm{z}_{0,t}}
\bigl\|\Phi_\text{HP}(\bm{y}) - \Phi_\text{HP}(\mathcal{D}(\bm{z}_{0,t}))\bigr\|_2^2,
\label{eq:latent_HGS_modification}
\end{equation}
thereby enabling a fair comparison of high-frequency preservation across both methods.

\begin{figure}[t]
    \centering
    \includegraphics[width=0.36\textwidth]{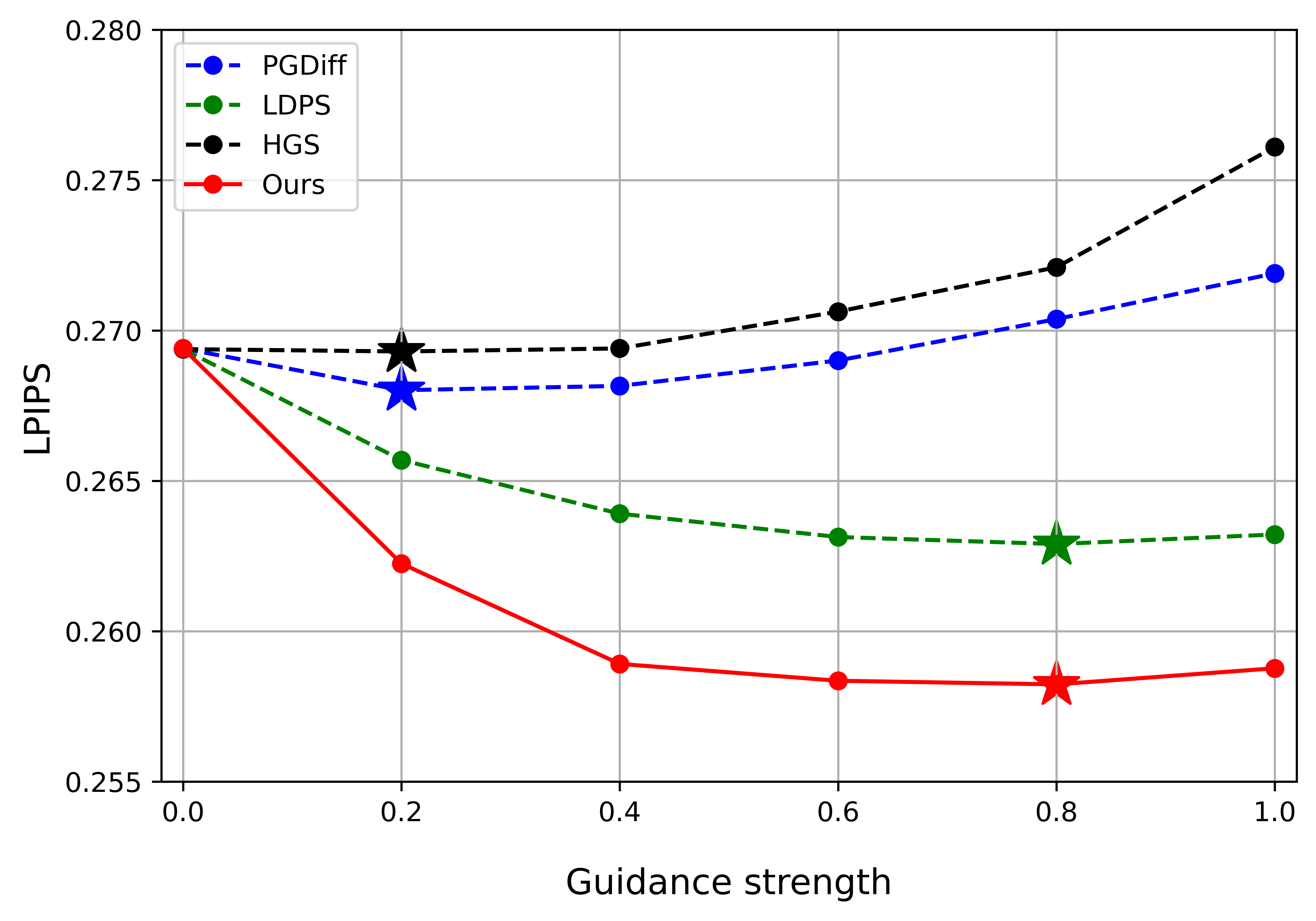}
    \vspace{-5pt}
\caption{Comparison of LPIPS performance among various guidance strategies applied to the shared baseline DiffBIR \cite{lin2023diffbir}. The horizontal axis indicates the normalized guidance strength (0–1), controlling the degree of guidance beyond the baseline (strength=0). Star symbols mark the best LPIPS points for each method. All results are evaluated on 4× blind super-resolution with medium degradation on the CelebA-HQ dataset.}
    \label{fig:pd_curve}
\end{figure}

As shown in Figure \ref{fig:pd_curve}, we compare our method with the above three inverse problem solvers: PGDiff \cite{yang2023pgdiff}, LDPS \cite{psld}, and HGS \cite{promptfix}.
All methods in this study share a common baseline, the pretrained LDM trained with DiffBIR \cite{lin2023diffbir}, resulting in identical starting points for all tradeoff curves.
Since varying guidance strengths lead to different reconstruction outcomes, the corresponding points are plotted to compare their respective curves, with star symbols indicating the best LPIPS points for each method.
Our proposed approach demonstrates a clear advantage by achieving the lowest LPIPS values compared with other methods.

\begin{figure}[t]
    \centering
    \includegraphics[width=0.36\textwidth]{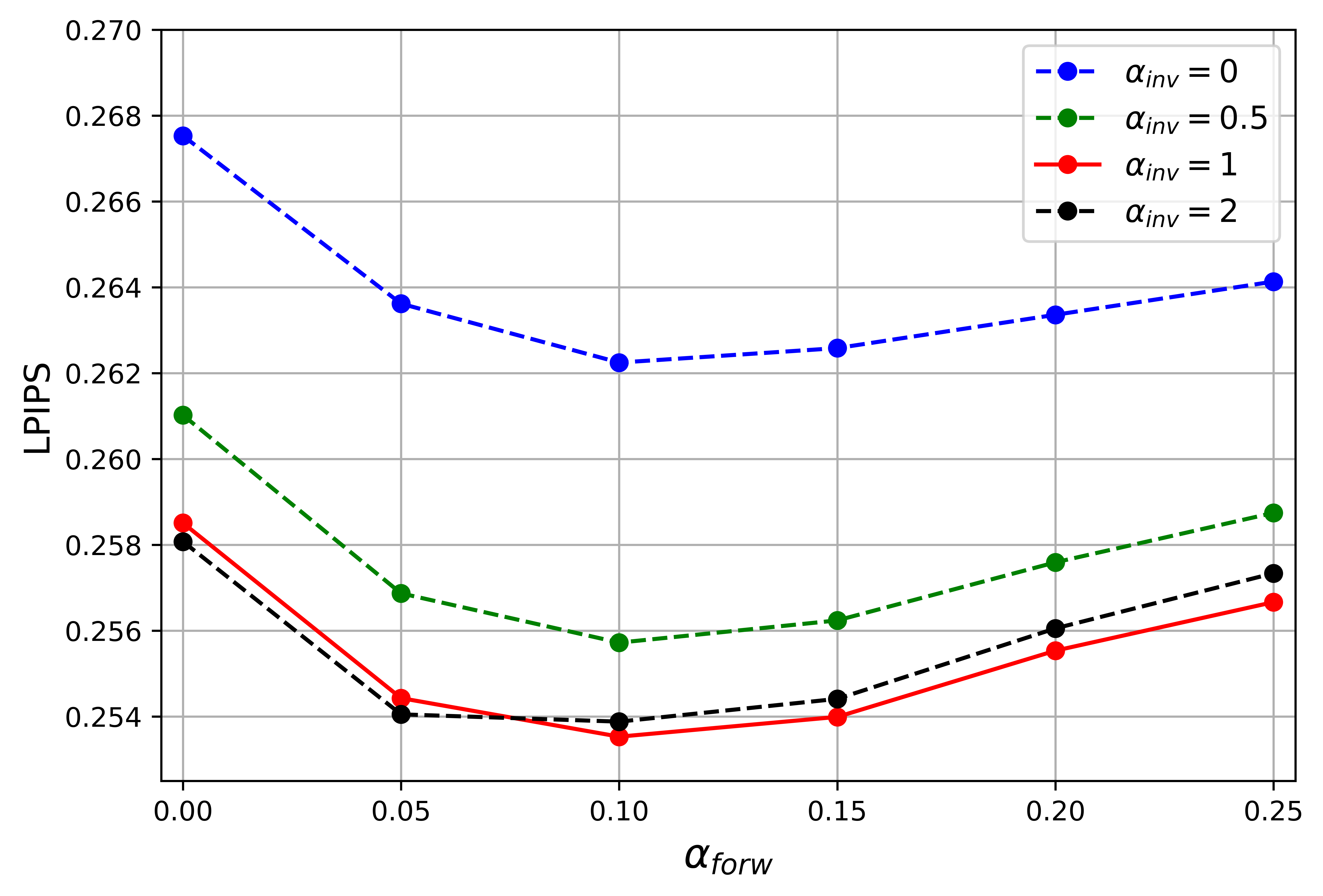}
        \vspace{-7pt}
\caption{Comparison of LPIPS across different $\alpha_{forw}$ and $\alpha_{inv}$ values as in line 8 of Algorithm \ref{alg:indigo_latent}.
Specifically, the blue line represents the case without applying our inverse INN guidance, while the red line corresponds to our default choice for $\alpha_{inv}$. The evaluation is conducted on 4x blind super-resolution with medium degradation on the CelebA-HQ dataset.
}
    \label{fig:alpha_forw_inv}
    \vspace{-0.5cm}
\end{figure}

\subsubsection{Ablation Study on
LatentINDIGO-PixelINN}
To demonstrate the effectiveness of our proposed guidance, we conduct an ablation study, as summarized in Table \ref{table_ablation_study}. Case 1 represents the baseline DiffBIR \cite{lin2023diffbir}, which utilizes their pre-trained $\bm{\epsilon}_\theta(\bm{z}_t, t, \bm{y})$ without any guidance. Cases 2-4 assess different components of our approach, with Case 4 serving as the default setting for our proposed LatentINDIGO-PixelINN.
Additionally, to explore further potential improvements, we evaluate PGDiff and HGS (as discussed in Section \ref{compare_guidance}), respectively.
When comparing Cases 1, 2, and 3, we can see that both {$\ell_{\mathrm{forw}}=\|{\bm{x}_{c,t}- \bm{y}}\|_2^2 $} and $\ell_{\mathrm{inv}}=
 \|{\varphi({\bm{x}}_{inv,t})- \varphi(\mathcal{D}(\bm{z}_{0,t}))}\|_2^2$ clearly contribute to performance gains on both PSNR and LPIPS.
We also observe that our regularization provides additional improvements (cases 3 and 4) by constraining latent representations to the manifold of real data.
Furthermore, Case 5 reveals that adding PGDiff guidance \cite{yang2023pgdiff} provides no further improvements in reconstruction performance. A comparison between Case 4 and 6 shows that although HGS guidance \cite{promptfix} yields a minor 0.01 PSNR improvement, it adversely affects LPIPS. Accordingly, these additional guidance terms are excluded from our default configuration.

\begin{table*}[]
\captionsetup{}
    \caption{{Ablation Study on LatentINDIGO-PixelINN on 4x {blind} SR with medium degradation on CelebA-HQ.}}
      \vspace{-0.2cm}
    \label{tab:as_blind}
    \centering
    \resizebox{0.55\textwidth}{!}{        
    \begin{tabular}{c|c|c|c|c|c||c|c}
\hline 
Case  &$\ell_{\mathrm{forw}}$&$\ell_{\mathrm{inv}}$ 
&Regularization &PGDiff \cite{yang2023pgdiff} &HGS \cite{promptfix} & PSNR $\uparrow$& LPIPS $\downarrow$    \\ \cline{1-8} 
1&--&--  &-- &-- &--&{24.72}&{0.2694}   \\ 
2&\ding{51}
&--  &-- &-- &--&{{25.17}}&{0.2636}   \\ 
3&\ding{51} &\ding{51}  &--  &--&--&{24.94}&{0.2605}   \\ 

\textbf{4}&\ding{51} &\ding{51}  &\ding{51}  &\textbf{--}&\textbf{--}&{25.43}&\textbf{{0.2535}}   \\ \hline
5&\ding{51} &\ding{51}  &\ding{51}  &\ding{51} &--&{25.27}&{0.2582}   \\ 
6&\ding{51} &\ding{51}  &\ding{51}  &-- &\ding{51}&\textbf{{25.44}}&{0.2537}   \\ 
\hline  
    \end{tabular}
}
    \label{table_ablation_study}
    \vspace{-0.3cm}
\end{table*}

\begin{table}[]
\captionsetup{}
    \caption{{Ablation study on various regularization strategies of LatentINDIGO. The experiment is conducted on 4x {blind} SR with medium degradation on CelebA-HQ.}}
    \vspace{-0.2cm}
    \label{tab:as_blind}
    \centering
\resizebox{0.48\textwidth}{!}{       
    \begin{tabular}{c|c|c}
\hline 
Case   & PSNR $\uparrow$& LPIPS $\downarrow$  \\  \hline
Ours w/o Regularization &24.94&0.2605 \\

\textbf{w/ Regularization in the first 15 steps (default) }&{25.43}&\textbf{0.2535}	   \\ 
{w/ Regularization in the first 30 steps  }&\textbf{25.60} &0.2694   \\ 
{w/ Regularization in the first 40 steps }&25.56 &0.2708 \\ 
w/ Regularization throughout all 50 steps &25.10 &0.3012	   \\ 
w/ Regularization every 5 steps &24.99 &0.2626	  	   \\

\hline  
    \end{tabular}
}
    \label{regu}
\end{table}
\begin{table}[]
\captionsetup{}

    \caption{{Ablation study on various training strategies of LatentINN. The experiment is conducted on 4x {blind} SR with medium degradation on CelebA-HQ.}}
        \vspace{-0.2cm}
    \label{tab:as_blind}
    \centering
    \resizebox{0.35\textwidth}{!}{
    \begin{tabular}{c||c|c|c|c}
\hline 
Case   & PSNR $\uparrow$& LPIPS $\downarrow$ & DISTS $\downarrow$  & FID $\downarrow$   \\  \hline
Default &25.07&\textbf{0.2490}&	\textbf{0.1570}&\textbf{22.89}  \\
1&\textbf{25.08	}&0.2512	&0.1591&22.97   \\ 
2&\textbf{25.08}&	0.2497&	0.1582	& 22.99  \\ 
3&\textbf{25.08}&	0.2493&	0.1582&	22.97 \\ 

\hline  
    \end{tabular}
    }
    \label{table_latentinn_loss}
    \vspace{-0.5cm}
\end{table}

\textbf{Analysis of $\alpha_{forw}$ and $\alpha_{inv}$:} One of the advantages of using INNs is that we can have a forward loss (line 6 in Algorithm \ref{alg:indigo_latent}) and a `backprojection' loss in image domain (line 7 in Algorithm \ref{alg:indigo_latent}). This increases the stability of the method and is crucial in latent diffusion posterior sampling.
This is shown in Fig. \ref{fig:alpha_forw_inv}, where we compare the LPIPS score across different $\alpha_{forw}$ and $\alpha_{inv}$ values (line 8 in Algorithm \ref{alg:indigo_latent}).
The blue curve corresponds to the scenario without applying the $\ell_{\mathrm{inv}}$ guidance, while the red line corresponds to our default choice for $\alpha_{inv}$. It can be seen that the blue line ($\alpha_{inv}=0$) performs significantly worse than other curves. Moreover, the point with $\alpha_{forw}=0.1$ and $\alpha_{inv}=1$ achieve the best LPIPS. Thus, these values are set as the defaults in Algorithm~\ref{alg:indigo_latent}.

\textbf{Discussion on Regularization:}
Among LDM-based IR methods, PSLD \cite{psld} applies regularization at every diffusion sampling step to keep samples on the real data manifold, whereas P2L \cite{p2l} does so intermittently to accelerate inference. To systematically explore the timing and frequency of regularization within an LDM-based IR framework, we conducted a set of more detailed experiments.

As shown in Table \ref{regu}, we start with our LatentINDIGO-PixelINN without any regularization and compare it with various alternatives. Specifically, we experiment with imposing regularization only during the first 15, 30, or 40 steps of the 50-step diffusion process, as well as throughout all 50 steps or every 5 steps. The results indicate that applying regularization in the early stages is generally beneficial, effectively suppressing off-manifold artifacts. For instance, focusing on the first 15 steps (our default setting) yields a notable improvement over the baseline in terms of both PSNR and LPIPS.
However, when regularization is extended to later stages, although PSNR may increase, LPIPS also becomes higher, indicating that excessive regularization can over-smooth and degrade fine-grained details generated by LDMs.

\begin{figure}
    \centering
    \includegraphics[width=0.7\linewidth]{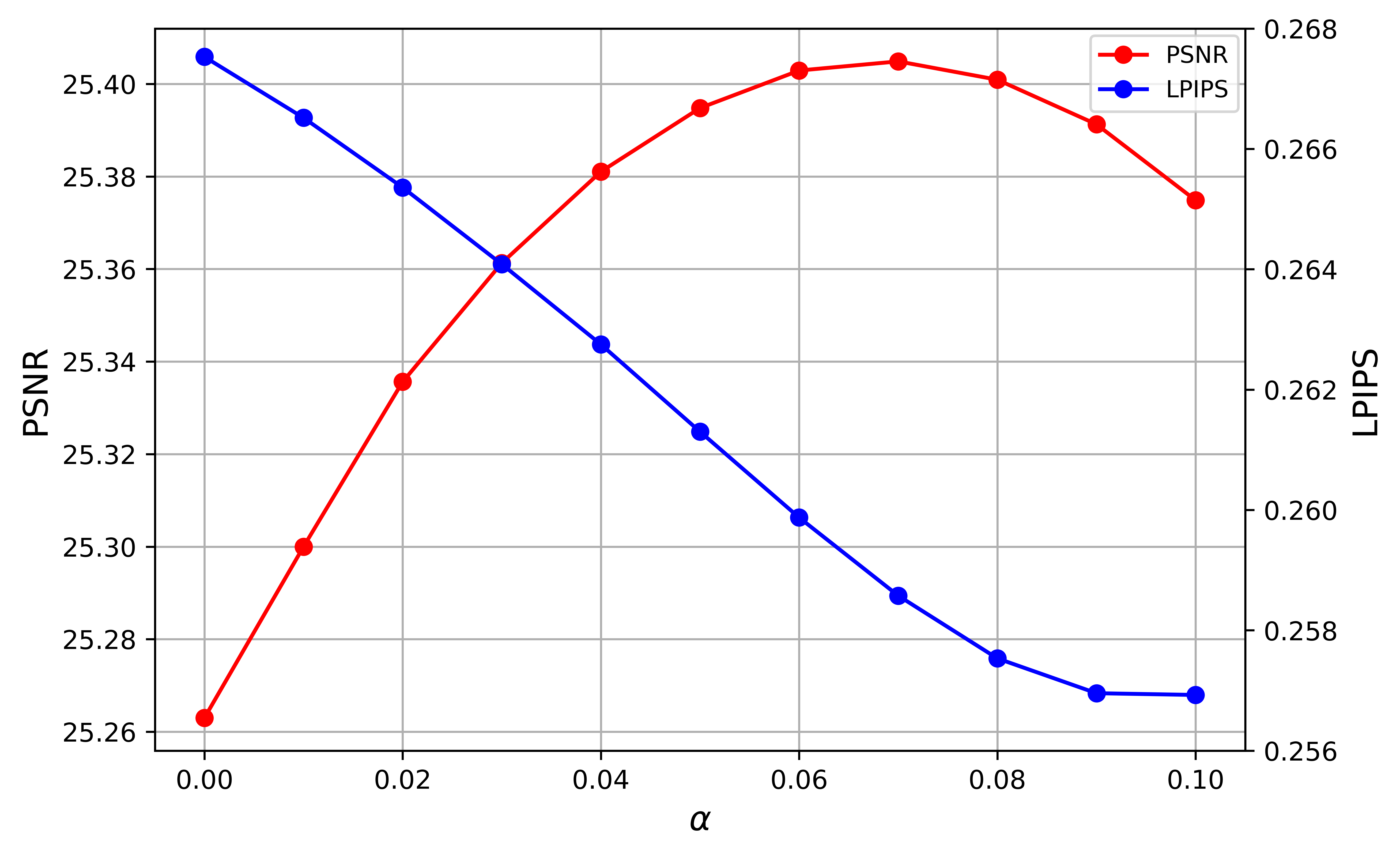}
            \vspace{-5pt}
   \caption{Comparison of PSNR (red line) and LPIPS (blue line) across different $\alpha$ values for our LatentINDIGO-LatentINN, corresponding to the update in line 6 of Algorithm \ref{alg:latent}. The evaluation is conducted on a 4× blind super-resolution task with medium degradation using the CelebA-HQ dataset. Here, the case with $\alpha =0$ represents our baseline DiffBIR\cite{lin2023diffbir}.}
    \label{fig:alpha}
\end{figure}

\begin{figure}[!tp]\footnotesize %\small
	\centering
\hspace{-0.2cm}
\begin{tabular}{c@{\extracolsep{0.05em}}c@{\extracolsep{0.05em}}c@{\extracolsep{0.05em}}c@{\extracolsep{0.05em}}c@{\extracolsep{0.05em}}c}
    
                &\includegraphics[width=0.092\textwidth]{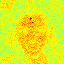}
    &\includegraphics[width=0.092\textwidth]{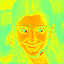}
    &\includegraphics[width=0.092\textwidth]{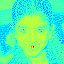}
    &\includegraphics[width=0.092\textwidth]{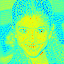}
        &\includegraphics[width=0.092\textwidth]{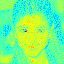}
\\
&Input & SwinIR &DiffBIR& Ours  & GT\\
    
	\end{tabular}
    % \vspace{-0.09m}
	\caption{{Visualization of the latent representation (first channel) of the results produced by different methods. Our proposed approach exhibits the closest alignment with the ground truth.}} 
 \vspace{-0.4cm}
	\label{fig:latent_comparison}
\end{figure}
\subsubsection{Ablation Study on LatentINDIGO-LatentINN}
\textbf{Training strategy of LatentINN:}
\label{inn_training}
In Section~\ref{latentinn_training}, we introduced the loss function for training our LatentINN, which comprises two components, denoted $\mathcal{L}_{forw}$ and $\mathcal{L}_{inv}$. In addition to these, we also experimented with the total variation (TV) loss, $ \mathcal{L}_{tv}\left ( {\mathbf{\Theta}_{lat}}  \right )=\frac{1}{N}\sum_{i=1}^{N} TV (\mathcal{D}( \bm{z}_{inv}))) $, given the need for the generated $z_{inv}$ to lie on the data manifold.
As shown in Table \ref{table_latentinn_loss}, in addition to our default model (trained with both $\mathcal{L}_{forw}$ and $\mathcal{L}_{inv}$), we evaluated LatentINN trained with only $\mathcal{L}_{forw}$ (Case 1), LatentINN trained with loss $\mathcal{L}_{forw}$ and $\mathcal{L}_{tv}$ (Case 2), and with all three losses $\mathcal{L}_{forw}$, $\mathcal{L}_{inv}$, $\mathcal{L}_{tv}$ (Case 3). These experiments allow us to assess the contribution of each loss component to the overall performance. We observe that the default setting achieves the best overall performance. In contrast, using only $\mathcal{L}_{forw}$ (Case 1) adversely affects the perceptual metrics LPIPS, DISTS, and FID, suggesting that $\mathcal{L}_{forw}$ does not sufficiently enforce the structural constraints on $\mathcal{D}( \bm{z}_{inv})$.
Incorporating the TV loss in Case 2 alleviates this issue, whereas further adding loss $\mathcal{L}_{inv}$ (Case 3) does not yield further improvements. Consequently, we adopt the combination of $\mathcal{L}_{forw}$ and $\mathcal{L}_{inv}$ as the default setting.

\textbf{Effect of LatentINN Guidance:} As shown in the sixth line of Algorithm~\ref{alg:latent}, the hyperparameter $\alpha$ is used to control the strength of the guidance toward the INN-optimized latent.
Figure~\ref{fig:alpha} illustrates the effect of different $\alpha$ values for our LatentINDIGO-LatentINN.
From the comparison of PSNR (red line) and LPIPS (blue line) across different $\alpha$ values, one can observe that smaller values of $\alpha$ lead to improved performance for both metrics, up to $\alpha = 0.07$ for PSNR and $\alpha = 0.09$ for LPIPS. Therefore, we set $\alpha = 0.08$ as the default value in our experiments. Although we also explored the gradient-based guidance in the latent space (as in Algorithm \ref{alg:indigo_latent}), it offered no noticeable advantage over this straightforward interpolation, thus reinforcing our choice of a simple yet effective linear blending scheme. To further illustrate how our guidance operates within the latent domain,  we visualize the latent representations obtained from the reconstruction results of various methods for the 4$\times$ blind SR task with medium degradation. As shown in Fig.~\ref{fig:latent_comparison}, our proposed approach achieves a closer alignment with the ground-truth latent, indicating that our guidance effectively integrates data fidelity and the diffusion prior, thus yielding more visually coherent reconstructions.
Finally, Table \ref{Tab:runtime} compares the runtimes of LatentINDIGO-PixelINN and LatentINDIGO-LatentINN, demonstrating the superior computational efficiency of the latter, which performs guidance entirely in the latent space.

\begin{table} \footnotesize
  \centering
  \caption{{Runtime comparison of LatentINDIGO-PixelINN and LatentINDIGO-LatentINN (based on DiffBIR, NFE = 50) computed with a single RTX 4090 GPU.}}
  \vspace{-0.1cm}
\resizebox{0.3\textwidth}{!}{
      \begin{tabular}{c|c c cc}
\hline
       Methods & Runtime (Seconds) \\
\hline
         LatentINDIGO-PixelINN  & {18.37}\\
          LatentINDIGO-LatentINN &\textbf{9.76}  \\

\hline
      \end{tabular}
    }

  \label{Tab:runtime}
  \vspace{-0.4cm}
\end{table}

\vspace{-0.2cm}
\section{Conclusion}
In this paper, we introduced a novel framework for blind image restoration (BIR) that leverages latent diffusion models (LDMs) and wavelet-inspired invertible neural networks (INNs). Unlike approaches that depend on predefined degradation operators, our method can handle any degradation by simulating it through the forward transform of the INN and reconstructs lost details via the inverse transform. We developed two variants, LatentINDIGO-PixelINN and LatentINDIGO-LatentINN, with the latter operating fully in the latent space to reduce computational complexity.
Both variants alternate between updating intermediate images with INN guidance and refining the invertible network parameters for unknown degradations, enhancing adaptability to real-world scenarios.
Our framework integrates with existing LDM pipelines without requiring additional retraining or finetuning and numerical results demonstrate that the proposed approach consistently delivers strong performance in terms of reconstruction accuracy and perceptual fidelity.
\vspace{-0.1cm}

\bibliographystyle{IEEEtran}
\bibliography{ref}
\end{document}